\documentclass{article}

\usepackage{graphicx} % Required for inserting images
\usepackage{amsmath}
\usepackage{amsthm}
\usepackage{amsfonts}
\usepackage{subfig}
\usepackage{natbib}
\usepackage{url}
\usepackage[ruled,linesnumbered]{algorithm2e}
\usepackage{multirow}
\usepackage{adjustbox}
\usepackage{tikz}
\usepackage{enumitem}
\usepackage{booktabs}
\usepackage[accepted]{icml2024}
 
\newcommand\todo[1]{\textcolor{red}{TODO: #1}}
\usepackage{appendix}

\icmltitlerunning{LLMs with  Episodic Memory Control}
 
\begin{document}

\twocolumn[
% \icmltitle{Larimar: LLMs with External Episodic Memory Control}
\icmltitle{Larimar: Large Language Models with  Episodic Memory Control}



\icmlsetsymbol{equal}{*}

\begin{icmlauthorlist}
\icmlauthor{Payel Das} {equal,yyy}
\icmlauthor{Subhajit Chaudhury}{equal,yyy}
\icmlauthor{ Elliot Nelson} {yyy}
\icmlauthor{Igor Melnyk}{yyy}%{comp}

\icmlauthor{Sarathkrishna Swaminathan}{yyy}
\icmlauthor{Sihui Dai}{yyy,comp}
\icmlauthor{Aur\'elie Lozano}{yyy} %{sch,yyy,comp}
\icmlauthor{Georgios Kollias}{yyy}%{comp}
\icmlauthor{Vijil Chenthamarakshan}{yyy}%{sch}
\icmlauthor{Ji\v{r}\'\i\, Navr\'atil}{yyy}%{sch}
\icmlauthor{Soham Dan}{yyy} %{yyy,comp}
\icmlauthor{ Pin-Yu Chen}{yyy}%{sch}
\end{icmlauthorlist}
% \author{Payel Das*, Subhajit Chaudhury*, Elliot Nelson, Igor Melnyk,\\ Sarath Swaminathan, Sihui Dai#, Aur\'elie Lozano, Georgios Kollias, Vijil Chenthamarakshan, \\ Ji\v{r}\'\i\, Navr\'atil, Soham Dan, Pin-Yu Chen}
\icmlaffiliation{yyy}{IBM AI Research}
\icmlaffiliation{comp}{Princeton University; work done during internship at IBM Research}

\icmlcorrespondingauthor{Payel Das and Subhajit Chaudhury}{daspa@us.ibm.com; subhajit@ibm.com}
% \footnote{Sihui Dai was an intern at IBM during this research.}

%\icmlkeywords{Machine Learning, ICML}
\vskip 0.3in
]
% \printAffiliationsAndNotice{\icmlEqualContribution\icml}
\printAffiliationsAndNotice{\icmlEqualContribution}
%\printNotice{\icmlEqualContribution}
\begin{abstract}
Efficient and accurate updating of knowledge stored in Large Language Models (LLMs) is one of the most pressing research challenges today. This paper presents Larimar - a novel, brain-inspired architecture for enhancing LLMs with a distributed episodic memory. Larimar's memory allows for dynamic, one-shot updates of knowledge without the need for computationally expensive re-training or fine-tuning. Experimental results on multiple fact editing benchmarks demonstrate that Larimar attains accuracy comparable to most competitive baselines, even in the challenging sequential editing setup,  but also excels in speed---yielding speed-ups of 8-10x depending on the base LLM ---as well as flexibility due to the proposed architecture being simple, LLM-agnostic, and hence general. We further provide  mechanisms for selective fact forgetting, information leakage prevention, and input context length generalization with Larimar and show their effectiveness. Our code is available at \texttt{https://github.com/IBM/larimar}.
\end{abstract}

\section{Introduction}

 Pre-trained Large Language Models (LLMs) have achieved impressive performance on various Natural Language Processing (NLP) tasks ~\cite{devlin2018bert,raffel2020exploring,brown2020language,vaswani2017attention}, and are often considered as knowledge repositories \cite{petroni2019language}. %However, accessing and modifying such knowledge in the LLM weights, although important for controllability, remains to be a challenging task~\cite{}.
 %There remain challenges to be resolved 
 In order to keep these models fact-relevant, safe, and ethical after deployment - the \emph{knowledge of the LLM needs to be constantly updated}. Thus, it is critical to develop efficient mechanisms to quickly update LLMs %for knowledge editing and forgetting  
 so that models can protect privacy, eliminate bias and hallucination, and catch up with new facts. 
 % Roughly speaking, when an LLM generates text composed of undesired, incorrect, or obsolete facts, model editing should remove the wrong fact from the LLM's ``memory", and optionally replace it with the desired outcome.  
Model editing should remove the undesired, incorrect, or obsolete facts from the LLM's ``memory", and optionally replace it with the desired outcome.  
 Similarly, the ability to quickly update the LLM can also help with the challenging problem of input \emph{context length generalization} beyond the training distribution,% i.e.,   extrapolating from short problem instances to longer ones,
 which is crucial when learning from datasets where longer context instances are rare \cite{anil2022exploring, kazemnejad2023impact}. %One evident way to tackle this issue is to scaling up the input context length. However, recent works show that interaction between LLM’s world knowledge, embedded in its parameters, and the factual information presented in the context lacks robustness and controllability \cite{li2022large, liu2023lost}.  An alternative path is to provide an LM decoder access to (key, value) pairs of previously seen subsequences from a non-differentiable memory while training the decoder on long-form text datasets, as proposed by  \cite{wu2022memorizing, wang2023augmenting}.   %augments the decoder with a non-differentiable memory that can store a large number of key-value pairs, query it,  and fuse retrieved key-value pairs from this module together with the in-context tokens for language modeling.  % One  prevalent challenge is generalization to long-form information beyond   the input length limit.  Another important challenge is  editing and unlearning of LLM's knowledge, %knowledge update 
 A straightforward solution is to fine-tune the model on the corrected/new datasets. Such an approach suffers the risk of overfitting and catastrophic forgetting~\cite{kirkpatrick2017overcoming, zhu2020modifying}%( Zhu et al., 2020)
 , as the knowledge is implicitly and distributionally encoded across the LLM parameters. 
Several lines of research have proposed effective and precise LLM editing (for comprehensive surveys on LLM editing, see \cite{li2022large, liu2023lost, zhu2020modifying}), which includes training an external memory model or a  hypernetwork model to work alongside with the frozen LLM. Another popular approach is to locate the original fact within LLM features and then do a local parameter update. As shown in Table \ref{tab:compare_methods}, both lines of methods face scalability problems due to overfitting and the need for retraining or locating for new states, causing a slow-down in editing speed.  The high memory needs for storing numerous edits provide a further obstacle in terms of scaling to sequential and batch editing setups. These challenges hinder the application of updating large language models in real-world industrial settings. Further, handling fact editing and selective fact forgetting appear challenging within the same methodological framework even for  current state-of-the-art editing methods \cite{patil2023sensitive}, while both new information learning and old information forgetting  are intrinsically related to each other in  in brain \cite{dempsey2022regional, autore2023adaptive}.

Humans, in contrast, can very quickly perform knowledge updating and generalization, both of which conform to rapid learning after seeing the first relevant instance. In the brain, such rapid learning is thought to depend on the hippocampus and its capacity for episodic memory. %Further, a single learning system often fall insufficient. For example, 
Consistently, while both semantic and working memory systems  struggle with sequential decision making tasks, the episodic memory systems are found to be beneficial \cite{blundell2016modelfree, lengyel2007hippocampal}.  %In brain, there exists multiple distinct memory systems, whose outputs are seamlessly integrated  to determine final behavior. 
 The  complementary learning systems (CLS) theory \cite{kumaran2016learning} provides rationale for coupling complementary \emph{fast} (hippocampus) and \emph{slow} (neocortex) learning systems in brain, former learning from single instances while later modeling the input distribution. %In CLS theory,  an integrated learning system comprised of different dedicated subsystems outperforms either subsystem. E.g.,  
 The neocortex-hippocampus interactions in brain is known to promote adaptive behavior via memorization and generalization \cite{sun2023organizing}. Further, it is proposed that the memory consolidation  from hippocampus to neocortex is facilitated  through the activation synchronized with multiple exact or false replays of the encoded experience in hippocampus -- suggesting hippocampus  taking the form of a \emph{generative associative network} \cite{ramirez2013creating}.

 \begin{figure*}[!ht]
    \centering
    \includegraphics[origin=c,width=0.95\textwidth]{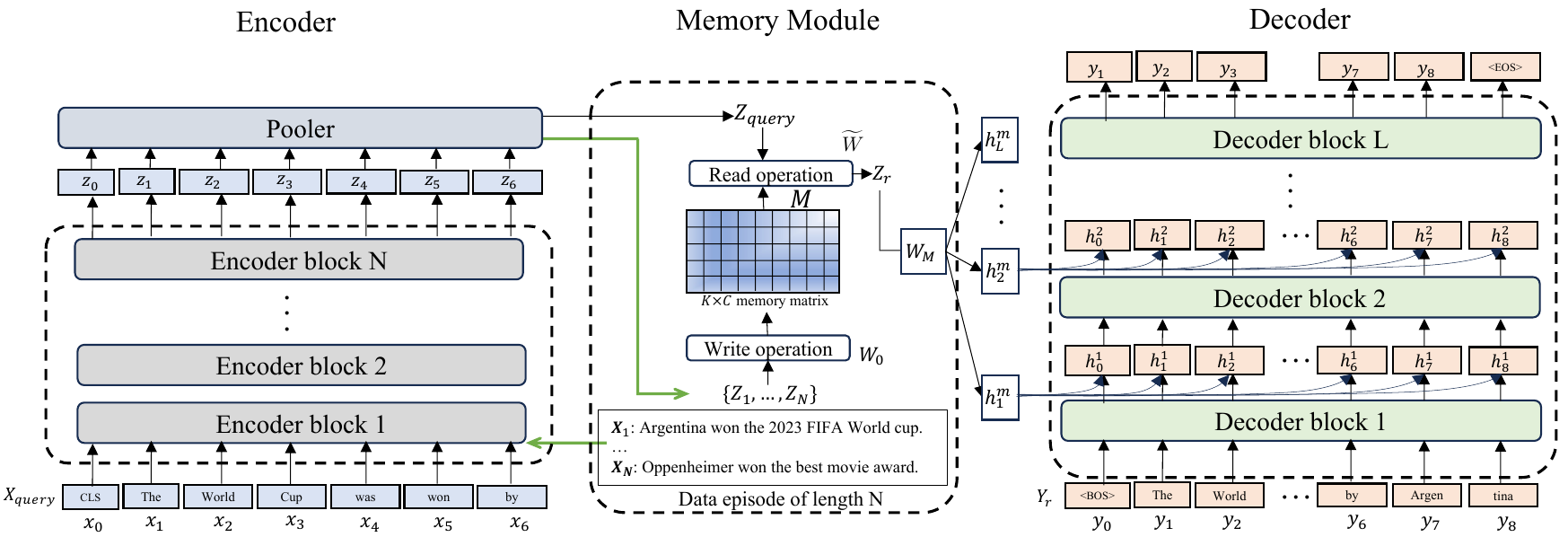}
       \vspace{-4mm}
    \caption{Larimar Architecture: $X$ and $X_{query}$ respectively denote data input and query, $Z$, $Z_{query}$ and $Z_r$ are the latent vectors, and $M$ is the fixed-size memory. $W$ and $W_0$ are reading/writing weights to memory. $W_M$ interfaces the readout from memory to the decoder. }
    \label{fig:Architecture}
\end{figure*}

Inspired by these insights, we propose \textbf{Larimar} -- a class of LLMs augmented with an external episodic memory controller. We follow the CLS view, where a hippocampal fast-learning system records samples as episodic memory, and a neocortical slow learning system (the LLM)  learns summary statistics of the input distribution as semantic memory. Our aim is to treat the episodic memory module as the global storage of the current set of factual updates or edits, and enforce this memory as a condition to the LLM decoder. It is important to learn to update this memory efficiently and accurately, without having to go through any training, as new edits arrive. %\SD{It's a bit confusing here since the experimental setup ends up using GPM instead of Kanerva Machine.  But the hierarchical Bayesian memory explanation is still relevant, the only difference if I recall correctly is that GPM uses deterministic $M$ }\AL{Modified accordingly}

%To tackle this, we seek to utilize the hierarchical Bayesian memory, referred as Kanerva Machine    %We seek to the sparse distributed memory (SDM) model proposed by Kanerva  as the model for the episodic memory, which  decouples the memory capacity  from the input dimensionality  by introducing a fixed-size distributed memory store (as opposed to slot-based memories) \cite{kanerva1988sparse}.  
%In order to comply to the hierarchical generative model of hippocampal formation \cite{stoianov2022hippocampal}, 
 %\cite{wu2018kanerva}.  
 %here,  a class of hierarchical conditional generative memory models is proposed, which is inspired by the sparse distributed memory (SDM) model  \cite{kanerva1988sparse}.  %Kanerva combining  slow-learning neural networks   with a fast-adapting memory \cite{wu2018kanerva}. 

To tackle this, we seek to utilize a hierarchical  memory, similar in spirit to the Kanerva Machine~\cite{wu2018kanerva}, where the memory writes and reads are interpreted as inference in a generative model. Specifically, we consider the memory model of~\cite{pham2021generative}, which 
treats the memory as deterministic, thereby
allowing reformulating the Bayesian updates of memory and address proposed in Kanerva Machine  as finding least-square solutions to linear systems.
 %The memory is treated as a global latent variable for an episode of input samples and refers to a distribution, $p(M)$, whereas the memory writes and reads are interpreted as inference in the generative model. 
Once updated, this fast-learning memory is then used to condition a slow-learning LLM decoder. %\VC{vision decoder?} %The memory refers to a distribution, $p(M)$. %which combines  slow-learning neural networks with a fast-adapting linear Gaussian model as memory. In this framework, named as Generative Kanerva Machines ,  memory writes and reads are interpreted as inference in a generative model, wherein memory is now treated as a distribution, $p(M)$. 
 The use of a \emph{global} memory associated a set of samples and the ability to \emph{fast write} to memory make this hierarchical  memory framework attractive for efficient LLM updating with respect to new knowledge. Implementation-wise, the memory is coupled to the LLM by end-to-end gradient descent on \emph{generic} data and does not assume access to edits.  During inference, the  new data is written to memory in one-shot, % memory is dynamically updated in onewith new data without undergoing any gradient-based training, 
 the updated memory then conditions the LLM decoding to enforce the edited output.   We further formalize  training-free  \emph{selective fact forgetting} and \textcolor{black}{\emph{information leakage prevention}} operations based on Larimar's  one-shot memory updating mechanism.  %enables Larimar to handle unseen edits in a training-free manner,  which is not the case for existing editing methods. 
 
To our knowledge, this is the first work that proposes  and demonstrates online distributed writing to a hierarchical conditional  memory model as a solution to test-time adaptation of LLMs to new knowledge.  
We demonstrate Larimar on single and sequential %and batch-based  
fact editing tasks on existing benchmarks and compared with baseline methods. Larimar provides accurate and precise editing across  these settings, while being \textcolor{black}{up to} 10 times faster compared to competitive model editing baselines 
%that are good at either sequential or batch editing setting (Table \ref{tab:compare_methods}). 
We further subject Larimar to  selective fact forgetting   \textcolor{black}{and information leakage prevention} and show its efficacy in those tasks.    %We  formalize a training-free  selective forgetting operation in Larimar, which is based on the one-shot memory updating mechanism. %We also show how Larimar's memory  enables   generalization to long-form information beyond   the input length limit. %which is important for long horizon planning and
% which is a challenge for current LLMs. 
%In contrast to prior non-differentiable memory-augmented transformers trained on long inputs, 
Lastly, we provide a simple recursive search-based solution that enables Larimar's  memory to  generalize to longer input context. %Larimar's   fixed-size, latent memory of  Larimar    can generalize to longer context by exploiting the dynamic memory update mechanism. 

\begin{table*}
\centering
\begin{adjustbox}{width=0.9\textwidth}
\begin{tabular}{lccccccc}
\toprule
{\textbf{Editor}} & {\textbf{+Edit Train}} & {\textbf{+Fact Trace}}  & {\textbf{Sequential Edit}} & {\textbf{Batch Edit}} & {\textbf{Forgetting\textcolor{black}{/Deletion} }}& {\textbf{Time (GPT-2)}} &  {\textbf{Time (GPT-J)}}\\
\midrule
%MEND (hypernetwork) & Yes & No &  3.9 & \\
ROME  & No & Yes & No & No & Yes & 4.8s & 13.9s \\
GRACE  & Yes & No &  Yes & No & No & 13.9s & 19.3s \\
Larimar & No & No &   Yes & Yes & Yes & 1.1s & 1.7s \\ \bottomrule
\end{tabular}
\end{adjustbox}
\caption{Requirement and capability comparison between Larimar, and two existing editing methods,   ROME and GRACE.}
\label{tab:compare_methods}
   \vspace{-2mm}
\end{table*}

Our contributions are: 
\begin{itemize}[leftmargin=*]
    \item Inspired by complementary learning mechanisms in the brain, we propose a class of episodic and adaptable memory-conditioned LLM architectures for  test time adaptation in real-time. Our method does not \emph{need} any time-intensive gradient-based learning or fact tracing within the LLM for performing the edit, providing a faster alternative for LLM updating. %spe\VC{add at test time?}. 
    \item We demonstrate the utility of this architecture on two relevant and challenging use cases: knowledge editing and input context length generalization. Larimar shows fast and accurate training-free adaptation to new inputs in both scenarios, compared to baseline editing methods and language models. 
    \item We show  selective fact forgetting and information leakage prevention  using one-shot memory updating.
    \item We provide a simple means to enable long context generalization in Larimar, based on a recursive search on its  memory space. 
    
    %\item We propose a LLM architectures with a separate explicit memory module - inspired by CLS and interactions between neocortex and hippocampus
   % \item associative memory
    %\item the memory-decoder coupling is trained such that decoder is controlled by the read/write operations corresponding to then memory
    %\item we explore both generative and deterministic memory modules
    %\item fast dynamic inference mechanism to update memory during test time -- which handles generalization better
    %\item introduce fact forgetting as a memory update operation 
    %\item speed of dynamic update
   % \item performance on new fact addition 
    %\item performance on existing fact editing benchmark 
    %\item performance on sequential and batch fact editing  
    %\item performance on fact forgetting 
    
\end{itemize}

%\begin{figure}[h]
 %   \centering
 %   \includegraphics[width=0.5\linewidth]{figures/ComplementaryLS.png}
  %  \caption{Complementary Learning System in the Human Brain}
   % \label{fig:CLS}
%\end{figure}

%Inspired by this..., we introduce Larimar -- generative LLM architectures that are coupled with generative associative memory modules.  

%In the framework, neocortex is approximated using generative LLM, whereas the memory unit represents hippocampus, former being much larger in size and complexity than later. 

%The memory unit is associative, generative, and sparsely distributed, while storing episodic memory. 

\section{Model Details}

\textbf{Notation}: 
We define input and output spaces as $\mathcal{X}$ and $\mathcal{Y}$, respectively. The model comprises an encoder $e: \mathcal{X} \to \mathbb{R}^C$ and a decoder $d: \mathbb{R}^C \to \mathcal{Y}$, linked via an adaptive memory. The encoder outputs in a latent space of dimension $C$. The memory uses $K$ rows to store encoded episodes of length $N$, with initial state $\mathbf{M}_0 \in \mathbb{R}^{K \times C}$ and updates through reading and writing weights $\mathbf{W}, \mathbf{W}_{0} \in \mathbb{R}^{N \times K}$, resulting in updated memory $\mathbf{M}$.
%We assume $\mathcal{X}$ is the input space and $\mathcal{Y}$ is the output space of the model. For the model we assume a pair of encoder and decoder coupled via an adaptive memory, $e: \mathcal{X} \to \mathbb{R}^C$ is the encoder and $d: \mathbb{R}^C \to \mathcal{Y}$ as the decoder. $C$ is the dimension of latent space, i.e., $e(X)$. For the memory module, $K$ is the number of rows of the memory matrix, and $N$ is the length of an episode written to memory. Using prior memory $M_0 \in \mathbb{R}^{K \times C}$, the encoding of an incoming episode  is written to memory.
%$W \in \mathbb{R}^{N \times K}$ is the reading weight matrix to memory and $W_{0} \in \mathbb{R}^{N \times K}$ is the writing weight matrix to memory. Result of the writing is the posterior memory $M$. 

% \begin{itemize}
%     \item $C$- dimension of latent space
%     \item $K$- number of rows of memory matrix
%     \item $T$- length of episode written to memory
%     \item $\mathcal{X}$ - input space
%     \item $\mathcal{Y}$ - output space
%     \item $M \in \mathbb{R}^{K \times C}$- posterior memory module
%     \item $W_{r} \in \mathbb{R}^{T \times K}$ - reading weight matrix to memory
%     \item $W_{w} \in \mathbb{R}^{T \times K}$ - writing weight matrix to memory%addressing latent variables of memory module
%     \item $e: \mathcal{X} \to \mathbb{R}^C$ - encoder
%     \item $d: \mathbb{R}^C \to \mathcal{Y}$ - decoder
% \end{itemize}

\textbf{Background Information:}
Generative memory networks are a type of associative memory networks that treat memory `read'/`write' and addressing operations as Bayesian inference where posteriors are updated when a new data episode arrives. Generative Pseudo-inverse Memory (GPM) framework proposed in~\cite{pham2021generative} reformulates these Bayesian updates as finding least-square solutions to linear systems, thereby enabling fast and efficient memory operations. One can then generate examples similar to a given input based on memory by sampling from the `read' distribution. In the following sections, we will elaborate the training and inference mechanisms, as we adapt that to train a LLM decoder. In Section \ref{sec:memory}, we will review the `write', `read' and `generate' operations derived in~\cite{pham2021generative} in detail, and formulate additional newly proposed `sequential writing' and `forgetting/unlearning' operations.

\subsection{Training} Given the memory $\mathbf{M}$,  Kanerva Machine aims %, Larimar architecture is composed of (i) an LM encoder $e$, (ii) a generative distributed memory matrix $M$, % of size $K \times C$, where $K$ is the number of memory slots and $C$ is the dimension of the encoding vectors, 
%and (iii) a pre-trained LM decoder $d$. 
%Tne encoder, the memory, and the decoder are trained end-to-end  and deco enable memory-conditioned decoding, these training objective is 
to maximize the conditional log-likelihood of $\ln p(\mathbf{X}|\mathbf{M})$, where $\mathbf{X}$ is an exchangeable (order invariant) episode: $\mathbf{X} = \{{x}_1,\ldots,{x}_N\}$,  a subset of the input data consisting of $N$ samples. A  variational lower bound of this conditional likelihood is optimized, similar to in variational autoencoders ~\cite{kingma2013auto}.
Consequently, the model learns to compress $\mathbf{X}$ in a memory $\mathbf{M}$, which then becomes a distributed associative memory. In practice, $\mathbf{M}$ is learned on a noisy version of the latent encodings $\mathbf{Z} + \xi$ where $\mathbf{Z} = e(\mathbf{X})$ for an episode. %consistent with Kanerva's memory model \cite{kanerva1988sparse}. 
In the remainder of this study, we use $\mathbf{M}$ as the posterior memory dependent on an episode $\mathbf{X}$, whereas $\mathbf{M}_0$ denotes a prior memory.
% corresponding to a different initial set of data samples. 
The reading weight matrix, $\mathbf{W}$, is a random variable to enforce generative ability of the model, for which we use a standard Gaussian prior $p(\mathbf{W}) \sim \mathcal{N} (0, I_{N \times K} )$ %the prior $p$ is initialized with a standard Gaussian $\mathcal{N} (0, I_{N \times K} )$
and posterior $q(\mathbf{W}) \sim \mathcal{N} (\overline{\mathbf{W}}, \sigma_{\mathbf{W}}^2\cdot I_{N \times K})$, where the mean $\overline{\mathbf{W}}$ is estimated from each episode and $\sigma_{\mathbf{W}}$ is learnable. The memory readouts are obtained as $\mathbf{Z}_{readout}={\mathbf{W}\mathbf{M}}$. 
The overall memory-augmented architecture is depicted in Figure~\ref{fig:Architecture}. 

%\subsection{Memory training}
 %\PD{start with conditional probability, factorize it. talk about weights, initialization}
During training all the three modules -- encoder ($e$), associative memory ($\mathbf{M}$), and decoder ($d$) -- are jointly trained and optimized for an episode $\mathbf{X}$, using the following loss: 
%\begin{multline}
%\label{eq:obj}
%    L = \mathbb{E}_{X \sim \text{data}} \big(\mathbb{E}_{q(W)} \ln p(X|W, M) + \\
%    \alpha \ln p(d(e(X)) 
%    - \beta D_{KL}(q(W)||p(W)) \\
%    - \mathbb{E}_{X \sim \text{pretrain}} \ln p(x_{i}|x_{i-1}..x_1) % + \delta \text{MSE}(e(X), \text{read}(e(X), \text{write}(e(X))) 
    %\big)
%\end{multline}.
\begin{align}
\label{eq:obj}
    L =& \mathbb{E}_{{\mathbf{X}} \sim \text{data}} \big(\mathbb{E}_{q({\mathbf{W}})} \ln p({\mathbf{X}}|\mathbf{W}, \mathbf{M}) \nonumber \\
     &+\alpha \ln p(d(e(\mathbf{X})) 
    - \beta D_{KL}(q(\mathbf{W})||p(\mathbf{W})) \nonumber\\
    &+ \mathbb{E}_{\mathbf{X} \sim \text{pretrain}} \ln p(\mathbf{x}_{i}|\mathbf{x}_{i-1}..\mathbf{x}_1). % + \delta \text{MSE}(e(X), \text{read}(e(X), \text{write}(e(X))) 
    %\big)
\end{align}
%\todo{align the equation}

The first term is the negative reconstruction loss with memory and $\mathbf{W}$, a $N \times K$ matrix. The second is the autoencoder's negative reconstruction loss without memory. The third is the KL divergence between prior $p(\mathbf{W})$ and posterior $q(\mathbf{W})$. To maintain decoder performance during training, a pretraining data regularization term is added.

%The first term is the negative reconstruction loss conditioned on  memory and the reading weights ${W}$, which is a $N \times K$ matrix.  The second term represents negative reconstruction loss for the autoencoder (without using memory).  The third term represents the KL divergence between the prior $p({W})$ and posterior $q({W})$ reading weights. To prevent possible performance degradation of the decoder during this generative memory augmentation training, a regularization term on pretraining data is further added. 

\subsection{Memory inference}
%Use of a linear gaussian model of memory allows the posterior memory to remain \SC{analytically tractable}. 
The memory at the end of training via backpropagation is then considered as a prior $\mathbf{M}_0$, the posterior memory $\mathbf{M}$ is updated in one-shot %using the Bayesian update (or equivalent) rule, enabling fast write when a new data episode arrives. To ensure efficient memory updating, we 
 by solving a minimization problem as proposed in \cite{pham2021generative}, which is 
%% [Transposition]
$\text{min}_{\mathbf{M}}||\mathbf{Z}_\zeta - \mathbf{W}_0 \mathbf{M}||_F^2$.
%% [ROME]
% $\text{min}_{\mathbf{M}}||{\mathbf{Z}_\zeta} - \mathbf{W}_0\mathbf{M}||_F^2$.
% that is equivalent to Bayesian update. 
This minimization problem, which corresponds to solving a linear system of equations, is efficiently done via computing matrix pseudo inverses.

\section{Memory operations}
\label{sec:memory}

\paragraph{Write, Read, Generate operations}
The three basic memory operations,  write in, read out, and generate, which  act upon the $\mathbf{Z}$ encodings, are cast as in \cite{pham2021generative}.  See Algorithm \ref{alg:memory_ops} for details.  %The output of the read and generate functions are converted into a text sequence through the decoder of the model.  
%We use the efficient one-shot memory update by solving a minimization problem, as proposed by ref \cite{pham2021generative}, that is equivalent to Bayesian update. This minimization problem, which corresponds to solving a linear system of equations, is efficiently done via computing matrix pseudoinverses. We note that pseudoinverses are approximated using the Ben-Cohen iterative algorithm \cite{ben1966iterative} \PD{number of approximation steps?}.

\paragraph{Sequential Writing and Forgetting}
%\todo{Georgios and Elliot: Add algorithms}

Given an initial set of encodings $\mathbf{Z}_0$ and writing weights $\mathbf{W}_0$, we initialize the memory matrix and key covariance matrix:
%% [Transposition]
\begin{align}
    \mathbf{M}_0 = \mathbf{W}_0^{\dagger} \mathbf{Z}_0, \quad \mathbf{C}_0 = \mathbf{W}_0^{\top} \mathbf{W}_0 
\end{align}
%% [ROME]
%\begin{align}
%    \mathbf{M}_0 = \mathbf{Z}_0 \mathbf{W}_0^{\dagger}, \quad \mathbf{C}_0 = \mathbf{W}_0 \mathbf{W}_0^{\top}
%\end{align}
To sequentially update the memory $\mathbf{M}_{i-1}$, either to add a new set of encodings $\mathbf{Z}_i$ or to forget a previously written set of encodings $\mathbf{Z}_i$, we jointly update the memory matrix and key covariance matrix for $i=1,2,...$ as follows:
%% [Transposition]
\begin{align}
    \mathbf{C}_i &= \mathbf{C}_{i-1} + \alpha_i \mathbf{W}_i^\top \mathbf{W}_i  \\
    \mathbf{M}_i &= \mathbf{M}_{i-1} + \alpha_i \mathbf{C}_i^{-1}  \mathbf{W}_i^\top (\mathbf{Z}_i - \mathbf{W}_i \mathbf{M}_{i-1})   
    %[ALTERNATE:] (C_{i-1} + \alpha_i W_i W_i^\top)^{-1}
    \label{eq:mem_update_seq}
 \end{align}
%% [ROME]
%\begin{align}
%    \mathbf{C}_i &= \mathbf{C}_{i-1} + \alpha_i \mathbf{W}_i \mathbf{W}_i^\top \\
%    \mathbf{M}_i &= \mathbf{M}_{i-1} + \alpha_i (\mathbf{Z}_i - \mathbf{M}_{i-1} \mathbf{W}_i) \mathbf{W}_i^\top
%    \mathbf{C}_i^{-1}
    %[ALTERNATE:] (C_{i-1} + \alpha_i W_i W_i^\top)^{-1}
%    \label{eq:mem_update_seq}
%\end{align}
When writing new encodings to memory, we use $\alpha_i=1$. 
When forgetting encodings which were previously written to memory with $\alpha_{i_{write}}=1$ at any $i_{write}<i$, we use $\alpha_i=-1$.
Eq. \eqref{eq:mem_update_seq} updates the memory sequentially such that it remains the least-squares solution for the growing sequence of data.
Assuming that $\mathbf{M}_{i-1}$ is the least-squares solution with respect to encodings $\mathbf{Z}_{0:i-1}$, that is,
%% [Transposition]
\begin{align}
    \mathbf{M}_{i-1} = \text{argmin}_{\mathbf{M}}\sum_{j=0}^{i-1}||\mathbf{Z}_j - \mathbf{W}_j \mathbf{M}||_2^2,
\end{align}
% [ROME]
%\begin{align}
%    \mathbf{M}_{i-1} = \text{argmin}_{\mathbf{M}}\sum_{j=0}^{i-1}||\mathbf{Z}_j - \mathbf{M}\mathbf{W}_j||_2^2,
%\end{align}
then Eq. \eqref{eq:mem_update_seq} with $\alpha_i=1$ ensures that $\mathbf{M}_i$ likewise is the least-squares solution with respect to $\mathbf{Z}_{0:i}$ (\cite{meng2023massediting}).
In the case $\alpha_i=-1$ and $\mathbf{Z}_i=\mathbf{Z}_{i_{forget}}$ for some $i_{forget}<i$, Eq. \eqref{eq:mem_update_seq} ensures that $\mathbf{M}_i$ is the least-squares solution with $\mathbf{Z}_{i_{forget}}$ removed from the data, that is,
%% [Transposition]
\begin{align}
    \mathbf{M}_i = \text{argmin}_{\mathbf{M}}\sum_{j=0,j\neq i_{forget}}^{i-1}||\mathbf{Z}_j - \mathbf{W}_j \mathbf{M}||_2^2,
\end{align}
%% [ROME]
%\begin{align}
%    \mathbf{M}_i = \text{argmin}_{\mathbf{M}}\sum_{j=0,j\neq i_{forget}}^{i-1}||\mathbf{Z}_j - \mathbf{M}\mathbf{W}_j||_2^2,
%\end{align}
%\JN{this is a meaty footnote with two equations being even numbered. Maybe consider it to be part of main text?}

%\EN{Static addressing mode:}
The weights can be computed either (following \cite{pham2021generative}) in terms of the current memory, $\mathbf{W}_i = \mathbf{Z}_i\mathbf{M}_{i-1}^{\dagger}$, or in terms of a fixed reference memory, $\mathbf{W}_i = \mathbf{Z}_i (\mathbf{M}^{(\rm ref)})^{\dagger}$. $\mathbf{M}^{(\rm ref)}$ remains unchanged across all sequential updates (i.e. is $i$-independent), is used only during inference, and can (optionally) be constructed using the episode of data encountered during inference.
%or, initialized from the trained memory matrix
%or, initialized randomly as long as its rows/columns overlap enough with the encodings encountered during inference
In the event that we wish to remove a given previously written encoding from memory, the fixed nature of $\mathbf{M}^{(\rm ref)}$ allows the original writing key $\mathbf{W}_{i_{write}}$ to be recomputed at a later point in the sequence $i_{forget}>i_{write}$, so that the information can be located in memory and removed.

\begin{algorithm}[!t]\footnotesize
\caption{Basic Memory operations \cite{pham2021generative}}\label{alg:memory_ops}

\DontPrintSemicolon
  \SetKwFunction{FWrite}{write}
  \SetKwFunction{FRead}{read}
  \SetKwFunction{FGenerate}{generate}
  \SetKwProg{Fn}{Function}{:}{}
  \Fn{\FWrite{$\mathbf{Z}$}}{ \tcp{$\mathbf{Z}$ - encoding of the episode to be written to memory (i.e. $\mathbf{Z} = e(\mathbf{X})$)}
        Sample $\xi \sim \mathcal{N}(0, \sigma_{\xi}^2I)$ \;\\
        Let $\mathbf{Z}_{\xi} = \mathbf{Z} + \xi$ \;\\
        Compute addressing weight $\mathbf{W}_0 = \mathbf{Z}_{\xi}\mathbf{M}_0^\dagger$ \;\\
        \tcp*{$\mathbf{M}_0$ is a learned parameter representing prior memory}
        Compute posterior memory $\mathbf{M} = \mathbf{W}_0^\dagger \mathbf{Z}_{\xi}$ \;\\
        \KwRet $\mathbf{M}$ \;
  }
  \;

  \Fn{\FRead{$\mathbf{Z}$, $\mathbf{M}$}}{
  \tcp{$\mathbf{M}$ - posterior memory from previous write}
  \tcp{$\mathbf{Z}$ - encoding of the read input (ie. $\mathbf{Z} = e(\mathbf{X})$)}
        Compute mean addressing weight $\overline{\mathbf{W}} = \mathbf{Z}\mathbf{M}^\dagger$ \;\\
        Sample $\mathbf{W} \sim \mathcal{N}(\overline{\mathbf{W}}, \sigma_{\mathbf{W}}^2I)$ \\ \tcp*{$\sigma_{\mathbf{W}}$ is a learned parameter}
        Compute output latent $\mathbf{Z}_{\text{read}} = \mathbf{WM}$ \;\\
        \KwRet $\mathbf{Z}_{\text{read}}$ \;
  }
  \;

  \Fn{\FGenerate{$\mathbf{M}$}}{
    \tcp{$\mathbf{M}$ is the posterior memory from a previous write}
        Sample $\mathbf{W} \sim \mathcal{N}(0, I)$\;\\
        Compute output latent $\mathbf{Z} = \mathbf{WM}$\;\\
        \KwRet $\mathbf{Z}$ \;
  }
  \;
  
\end{algorithm}

%For training, we initialize weights for the Optimus AE architecture from an Optimus AE model pretrained on English Wikipedia.  We train with batch size 256, AdamW optimizer, learning rate 0.00005, and linear learning rate scheduler.  We also anneal the strength of the KL term $\beta$ using the same beta annealing scheduler as used by \cite{li2020optimus} for training Optimus VAE.

%To model associative generative memory for text, we combine GPT with generative pseudoinverse memory (GPM) \cite{pham2021generative}.  The GPM provides 3 operations for memory that act upon the latent space of the autoencoder: writing to memory, reading from memory, and generating from memory.  Algorithm \ref{alg:memory_ops} specifies the steps within these operations.  The output of the read and generate functions can be converted into a text sequence through the decoder of the model.  We note that pseudoinverses are approximated using the Ben-Cohen iterative algorithm \cite{ben1966iterative}.
\section{Scope Detector}
We also optionally use a scope detection mechanism to detect if the incoming query is close to the facts written in the memory, which is conceptually similar to SERAC~\cite{mitchell2022memory}.  If the query is in-scope, then the corresponding readout from memory is passed to the decoder for memory-conditional decoding otherwise the query is subjected to unconditioned decoding. We consider two different scenarios:

\textbf{External encoding-based scope detector (ESD)}: Sample embeddings are estimated from an external sentence encoder (MiniLM\footnote{https://huggingface.co/sentence-transformers/all-MiniLM-L6-v2}) trained on ~1.1B sentence pairs and with an output space dimensionality of 384. The ESD stores encoded facts as vectors in its own scope storage. At test time, given an encoded input sentence, 1-nearest neighbor cosine similarity is calculated and serves as detection score. Any multi-sentence input is first split into isolated sentences, each of which is processed separately and maximum similarity is taken. Measured on 3800 positive and negative samples from the EasyEdit data set, this ESD model achieves a detection equal-error-rate of 2.9\% and an F1 score of 0.974. 

%\JN{Subhajit could you pls check this part?}
%\VC{xxx}
\textbf{Internal Encoding-based scope detector (ISD)}: Larimar encoder $e$ is used to embed CounterFact samples. The encodings are then used to train a binary scope classifier, where positive samples come from rephrasings of an original fact and negative data correspond to neighboring facts.  
%Note, we are using a common scope detector trained on counterfact across edit datasets to keep the overall framework dynamically adaptable to new knowledge, which is different from SERAC~\cite{mitchell2022memory} that uses a data-dependent scope detector. %\PD{CONFIRM}

\section{Results}
\label{sec:results}

\textbf{Implementation}: 
We employed a BERT large encoder~\cite{devlin2018bert} combined with either a GPT2-large~\cite{radford2019language} or a GPTJ-6B decoder and a memory matrix (512x768) for our training experiments, naming the resulting models Larimar-1.3B and Larimar-6B, respectively.
Our training data comprised 7.6 million examples constructed by splitting WikiText ~\cite{merity2016pointer} texts to small chunks of 64 tokens. We used existing pretrained weights from Huggingface to initialize the training. From the $Z_r$ (the readout vector), using learned linear projection $W_{M}$, the hidden states are transformed and broadcasted to act as a KV cache across all layers. During the decoder forward pass, this compressed KV cache is used as past KV cache values to generate the memory-controlled output. The hidden size for GPTJ, for example, is 4096 with 28 layers.
 In testing, the Larimar-1.3B model achieved a perplexity of 14.6, while the Larimar-6B model reached 15.9 on 1,000 random WikiText samples, indicating that adding memory barely affects performance. We trained Larimar-6B models for 10 epochs using Adam optimizer, learning rate 5e-6 and batch size 32. For the Larimar-6B's training, we used a setup with eight NVIDIA A100-80GB GPUs on a single node, utilizing bfloat16 precision and PyTorch Lightning with the DeepSpeed ZeRO Stage 2 for efficient distributed training.

\subsection{Wall Clock time}
Table~\ref{tab:compare_methods} presents the wall clock time for each editing method across 10 edits, calculated within the EasyEdit framework~\cite{yao2023editing} on a single A100 GPU. Results show that Larimar is 4-10x times faster compared to ROME \cite{meng2022locating} and GRACE \cite{hartvigsen2022aging}, two most competitive existing LLM editing baselines. Table \ref{tab:time} further provides a edit-time comparison within other existing baselines, as shown in \cite{yao2023editing}, establishing Larimar's advantage on high-speed editing. 
Table \ref{tab:compare_methods} further lists Larimar's abilities to handle edits in a training- or tracing- free manner, enabling high-speed editing,  handling selective forgetting, and maintain ability to  sequential editing setup.

%The wall clock time for each edit method when performing 10 edits is reported in Table~\ref{tab:compare_methods}. The time taken for baseline methods are estimated within the EasyEdit framework~\cite{yao2023editing}. All numbers are computed on a single A100 GPU. Results show that Larimar is 4-10x times faster compared to ROME and GRACE.

\begin{table}[!ht]
\centering
\begin{adjustbox}{width=0.48\textwidth}
\begin{tabular}{lrrrrrr}
\toprule
\multirow{3}{*}{\textbf{Editor}} & \multicolumn{2}{c}{\textbf{Edit Success}} & \multicolumn{2}{c}{\textbf{Paraphrase}} & \multicolumn{2}{c}{\textbf{Neighborhood}} \\
\cmidrule(lr){2-3}\cmidrule(rr){4-5}\cmidrule(lr){6-7} & S & M & S & M & S & M \\
\midrule
GPT-2 XL & 22.2 & -4.8 & 24.7  & -5.0  & 78.1 & 5.0 \\\midrule
FT & \textbf{100.0} & 98.8 & 87.9 & 46.6 & 40.4 & -6.2 \\
FT+L & 99.1 & 91.5 & 48.7 & 28.9 & 70.3 & 3.5 \\
KN & 28.7 & -3.4  & 28.0 & -3.3 & 72.9 & \textbf{3.7} \\
KE & 84.3  & 33.9 & 75.4 & 14.6 & 30.9 & -11.0 \\
KE-CF & \textbf{99.9} & 97.0 & 95.8 & 59.2 & 6.9 & -63.2 \\
MEND & 99.1 & 70.9 & 65.4 & 12.2 & 37.9 & -11.6 \\
MEND-CF & \textbf{100.0} & {99.2} & \textbf{97.0} & \textbf{65.6} & 5.5 & -69.9 \\
ROME & \textbf{100.0} & 97.9 & \textbf{96.4} & {62.7} & \textbf{75.4} & \textbf{4.2} \\
% \textbf{Larimar-1.3B} & \textbf{100.0} & \textbf{99.8} & 41.7 & 0.4 & {74.7} & 1.6 \\
\textbf{Larimar-1.3B} & \textbf{100.0} & \textbf{99.7} & 83.5 & 50.5 & \textbf{74.7} & 1.8 \\
\textbf{Larimar-1.3B} + 1 rephrase & \textbf{100.0} & \textbf{99.6} & 89.8 & 62.4 & {73.3} & 0.5 \\
\textbf{Larimar-1.3B} + 2 rephrases & \textbf{100.0} & \textbf{99.6} & 90.8 & \textbf{63.6} & 73.3 & 0.6 \\
\midrule\midrule
GPT-J & 16.3 & -7.2 & 18.6 & -7.4 & 83.0 & 7.3 \\\midrule
FT & \textbf{100.0} & \textbf{99.9} & {96.6} & \textbf{71.0} & 10.3 & -50.7 \\
FT+L & 99.6 & 95.0 & 47.9 & 30.4 & 78.6 & 6.8 \\
MEND & 97.4 & 71.5 & 53.6 & 11.0 & 53.9 & -6.0 \\
ROME & \textbf{99.9} & \textbf{99.4} & \textbf{99.1} & \textbf{74.1} & 78.9 & 5.2 \\
PROMPT & 99.7 & 80.9 & 91.0 & 32.9 & 37.9 & -2.8 \\
IKE (w/ 32 demonstrations)  & \textbf{100.0} & 91.7 & 95.2 & 64.5 & 77.0 & \textbf{35.2} \\
IKE (w/o paraphrases) & \textbf{100.0} & -- & 73.8 & -- &  \textbf{83.4} & -- \\
IKE (w/o neighbors) & \textbf{100.0} & -- & \textbf{99.8} & -- &  11.5 & -- \\
%\textbf{Larimar-6B} & 99.6 & 96.0 & 76.5 & 22.4 & \textbf{80.2} & 3.9 \\
\textbf{Larimar-6B} & 99.6 & 96.0 & 88.4 & 54.7 & \textbf{80.4} & 4.22 \\
\textbf{Larimar-6B} + 1 rephrase & 99.7 & 95.9 & 92.9 & 67.0 & 79.3 & 3.5 \\
\textbf{Larimar-6B} + 2 rephrases & 99.8 & 95.7 & 93.6 & 67.0 & {79.2} & 3.38 \\
 \bottomrule
\end{tabular}
\end{adjustbox}
\caption{Single fact editing on CounterFact dataset. Top two best editing methods  are highlighted.  Larimar uses dynamic memory updates with memory-conditioned decoding \textcolor{black}{and does not require gradient update on edit samples, as  opposed to methods needing training  (FT, FT+L, MEND) or tracing plus decoder updating (ROME) on edit samples (ROME) or  in-context demonstrations (IKE) of (paraphrased) edits and neighboring samples retrieved from a corpus. Though generalization increases when Larimar's memory is augmented with rephrases at test time. %the generalization performance is significantly enhanced.
}} %\textcolor{red}{Add *, triangle, etc to show methods that are doing fine-tuning vs tracing vs no Larimar, Add numbers from IKE without update and retain, drop PROMPT}
\label{tab:counterfact}
   \vspace{-2mm}
\end{table}
 %Upper panel corresponds to models built on top of GPT-2  while the lower portion is for GPT-J based models. 

\subsection{Single Fact editing}
\label{sec:cf_edit}
We compare the performance of Larimar against a number of recently proposed knowledge editing approaches on the CounterFact dataset \cite{meng2022locating} designed for testing language models handling of counterfactual edits. It includes 21,919 records to assess if the models can learn new facts rather than simply memorizing the target words. Following other works \cite{meng2022locating, zheng2023edit}, we used the first 2000 samples of this dataset and report the average over single fact editing  results  for Larimar-1.3B and Larimar-6B in Table \ref{tab:counterfact}. 
The  baseline performances are taken from \cite{meng2022locating, zheng2023edit} (see Related Work and Appendix for details on baseline methods). As opposed to training the LLM on edits, or causally tracing the original fact within LLM  and updating the relevant parameters to reflect edit, we leverage Larimar's one-shot memory update for editing. Wherein, the memory posterior is updated as the  edit(s) of interest is written, and then the updated memory is queried. The read-out from the memory then conditions the decoder to output the edit.   

The evaluation metrics used in Table \ref{tab:counterfact} are as follows:
% \begin{itemize}
\textit{Edit Success}, which is the percent of cases where the edited fact $(s, r, o^*)$, (subject, relation, object) with modified object has higher probability than the one based on the original object $(s, r, o^c)$. Specifically, column $S$ measures percentage of $\mathbb{P}[o^*] > \mathbb{P}[o^c]$ cases, while $M$ is the average of $\mathbb{P}[o^*] - \mathbb{P}[o^c]$ in the logits space of the language model. 
\textit{Paraphrase} measures the same performance on $(s, r, o^*)$ but using paraphrased prompts. \textit{Neighborhood} evaluates the model's ability to retain knowledge about the original object but in the context of neighboring subjects $s^\prime$: $(s^\prime, r, o^c)$. Here the column  $S$ reflects percentage of cases where $\mathbb{P}[o^c] > \mathbb{P}[o^*]$, while $M$ is the average $\mathbb{P}[o^c] - \mathbb{P}[o^*]$. 

As can be seen, when compared to existing editing baselines, Larimar achieves comparable performance in successfully editing new facts, and in the ability to handle neighborhood prompts. For example, when compared with ROME, Larimar performs on par when based on GPT-2 Large, and better when based on GPT-J on editing success and neighborhood specificity, while there remains room to improve generalization.  When compared to  existing in-context editing approaches (PROMPT and IKE) \cite{zheng2023edit}, \textcolor{black}{Larimar does not need multiple in-context demonstrations of the edits and its paraphrases, as well as of neighboring facts, to the decoder, which are retrieved from a corpus.  However, as shown in \textcolor{black}{Tables \ref{tab:counterfact} and \ref{tab:counter_ablate_para}}, when Larimar has access to one or two additional paraphrase(s) per fact, by writing it in the memory,  the generalization performance increases from \textcolor{black}{88.4 to 93.6}. Note that in this setup the average number of added paraphrase per fact is \textcolor{black}{at most two} and we queried the model with a paraphrased prompt unseen by the memory. %Also, such a setup is not easily amenable for editing in sequence or in batch, neither can handle knowledge forgetting. 
}  And, writing latent encodings of paraphrases in memory and conditioning the decoder on the memory is more cost-effective than in-context demonstrations, as the context length does not increase with Larimar's in-memory mechanisms. 
Ablation experiments in Appendix shows that a scope detector, either trained on Larimar encodings or encodings from an external LLM, helps with better paraphrase generalization, at the cost of sacrificing the   neighborhood specificity. In the absence of a scope detector, the same approach of augmenting memory with two additional rephrases provide an additional 2-3\% increase in generalization, % whereas neighborhood specificity is affected significantly, 
irrespective of the dataset (see next paragraph). Throughout the paper, Larimar is configured with a scope detector, unless otherwise mentioned. %For details, see Appendix.  
% \subsection{ZsRE Evaluation}
%We also assess Larimar on ZsRE benchmark \cite{zsre2017}, which is a question answering (QA) dataset used for relation extraction via reading comprehension. Table \ref{tab:zsre} shows the results. The performance scores for GPT-2 XL based baselines are from \cite{meng2022locating}, while for baselines based on GPT-J are from \cite{li2023pmet}. Note that in contrast to CounterFact evaluation, here the metrics are computed by counting number of exact matches of the top probability token with ground truth $\mathbb{I}[o^* = \text{argmax}_o\mathbb{P}[o]]$.
%Similar to the results on CounterFact, for ZsRE benchmark Larimar shows effectivity at both editing and paraphrasing tasks with similar or slightly lower performance in the Neighborhood category. It also shows consistency of performance whether it is based on GPT-2 or GPT-J, showing the model-agnostic editing efficiency of Larimar. 

We also evaluated Larimar on the ZsRE benchmark \cite{zsre2017}, a QA dataset for relation extraction through reading comprehension, with results displayed in Table \ref{tab:zsre}. Performance scores for GPT-2 XL based baselines are cited from \cite{meng2022locating}, whereas performance of ROME on GPT-J was independently estimated by us. % and \cite{li2023pmet}, respectively. 
Unlike the CounterFact evaluation, this assessment uses exact match counts for scoring $\mathbb{I}[o^* = \text{argmax}_o\mathbb{P}[o]]$. Compared to baselines, Larimar demonstrates effective editing and comparable neighborhood specificity  on ZsRE, with  slightly lower generalization, maintaining consistent results across GPT-2 and GPT-J decoders, underscoring its model-agnostic editing capabilities. We again see that test-time augmentation of the memory with additional paraphrases of the fact boosts generalization from 70.4\% to 82.2\%, with two rephrases written in memory (Table~\ref{tab:ke_gen_zSRE}).

%\begin{table}
%\centering
%\begin{adjustbox}{width=0.42\textwidth}
%\begin{tabular}{lrrrrrr}
%\toprule
%\multirow{3}{*}{\textbf{Editor}} & \multicolumn{2}{c}%{\textbf{Edit Success}} & \multicolumn{2}{c}%{\textbf{Paraphrase}} & \multicolumn{2}{c}%{\textbf{Neighborhood}} \\
%\cmidrule(lr){2-3}\cmidrule(rr){4-5}\cmidrule(lr){6-7} & S & M & S & M & S & M \\
%\midrule
%Larimar (ext) & 99.6 & 96.0 & 76.3 & 22.1 & 80.2 & 3.9 \\
%Larimar(int-ep4) & 99.8 & 89.8 & 83.5 & 23.5 & 82.1 & 6.1 \\
%Larimar(int-ep8) & 99.9 & 92.0 & 82.9 & 16.5 & 81.3 & 5.3 \\ %\bottomrule
%\end{tabular}
%\end{adjustbox}
%\caption{Ablation experiment with Larimar scope (trained on cf data).}
%\label{tab:counterfact_abl}
%\end{table}

\subsection{Sequential Fact Editing}
\label{sec:seq_edit}

We  evaluated Larimar on a sequential editing task, following the setup of \cite{hartvigsen2022aging}, which tackles the issue of forgetting previous edits after multiple sequential edits. Hartvigsen et. al. introduced a continual editing method that integrates an adaptor to update a codebook of edit key-value pairs with a pre-trained language model (GPT2-XL), showing memory retention during sequential edits. We adapt Larimar to this experimental setup, wherein  a subset of 200 facts with 5 rephrasings each is selected from the ZsRE validation dataset for testing. %wherein  a subset of 1000 sentences is selected from the ZsRE validation dataset for testing, which is the first 200 facts with at least 5 rephrasings each.
In Larimar, a sequential edit is handled by updating the global memory through Eq. \eqref{eq:mem_update_seq}, again requiring  no gradient-based update on incoming edits. For each edit, the encoding of the rephrased query concatenated with the corresponding   answer is written to memory. 
We assessed Larimar's performance, compared to GRACE, using the edit retention rate (ERR), which is the mean F1 score after 1000 sequential edits when querying the memory with the encoded query $\mathbf{Z}_{\rm query}$ for each written fact. Larimar is not finetuned on question-answer data; instead, we write each question-answer pair as a fact in the memory and query the memory with the original question. %We report on the Test Retention Rate (TRR) by evaluating Larimar decoder's perplexity on 1000 random test samples  from wikitext using a separate language model. In comparison, baseline models compute TRR from mean F1 scores from 1000 random samples of NQ data. 
Results show Larimar's comparable ERR performance to GRACE, that is the best-performing baseline, %, while preserving its original test set performance. 
while Larimar performing editing approximately 10 or more times faster than GRACE on GPT-2 XL. We found the batch editing method  SERAC  (referred as DEFER in 
\cite{hartvigsen2022aging};) to perform worse (ERR=0.31) compared to both   Larimar (ERR=0.98) and  GRACE (ERR=0.96) on  sequential editing on an edit dataset containing duplicates of fact and corresponding rephrasings,  though SERAC was trained on fact edits.

\begin{table}
    \centering
    \begin{adjustbox}{width=0.45\textwidth}    
    \begin{tabular}{c c c c c}
\toprule
{\textbf{}} & {\textbf{MEND}} & {\textbf{GRACE}} & {\textbf{Larimar-1.3B}}  & {\textbf{Larimar-6B}}\\
\midrule
%MEND (hypernetwork) & Yes & No &  3.9 & \\
Edit Retention Rate & 0.27 & \textbf{0.93} & \textbf{0.97} & 0.92\\
\bottomrule
\end{tabular}
\end{adjustbox}
\caption{Sequential editing on ZsRE, showing Larimar does not forget older edits, baselines  are from   \cite{hartvigsen2022aging}.
}
\label{tab:zsre_seq}
\vspace{-2mm}
\end{table}

%To compare to the generalization results of \cite{hartvigsen2022aging}, we use the same dataset comprising 1000 facts from ZsRE with 10 rephrasings each, split into edit and holdout sets. %\EN{ref appendix for details about basis memory} 
%In Figure \ref{fig:zsre_seq_gen} we show the mean F1 score on the holdout set %\EN{or, a size-XXX sample from the holdout set, averaged over XXX runs} 
%as a function of the number of sequential writes to memory made so far using the edit set, and compare to the results of \cite{hartvigsen2022aging}.\footnote{In Figure \ref{fig:zsre_seq_gen}, we show the best-performing curves from the holdout plots in Figure 4 of \cite{hartvigsen2022aging}.}
%Despite starting without prior exposure to the facts written to memory, Larimar is able to match the generalization performance of GRACE after $O(1000)$  writes to memory, or $O(1)$ write per fact, when the memory size $K$ exceeds the number of facts.

We also evaluated Larimar's  generalization to rephrased prompts, again comparing to GRACE.
We use (i) a dataset of unique 1000 ZsRE facts, each with 10 variations, divided into edit and holdout sets, and (ii) an edit/holdout dataset with more ($\approx20$ per fact rephrasings and fewer ($\approx500$) unique ZsRE facts. %Similar to \cite{hartvigsen2022aging}, we  use a  dataset of 1,000 ZsRE facts, each with 10 variations, divided into edit and holdout sets.
Our analysis, depicted in Figure \ref{fig:batch} (b), examines the mean F1 score on the holdout set against the number of memory writes using the edit set, {\color{black}compared to GRACE %\footnote{See \href{https://github.com/Thartvigsen/GRACE/tree/main/grace}{https://github.com/Thartvigsen/GRACE/tree/main/grace}} 
on the same datasets.\footnote{We use GRACE with $\epsilon_{\rm init}=3.0$ to edit block 4 of T5 \cite{hartvigsen2022aging}.}} %\footnote{In Figure \ref{fig:zsre_seq_gen}, we show the best-performing curves from the holdout plots in Figure 4 of \cite{hartvigsen2022aging}.} 
As Larimar has no knowledge of upcoming edits, it starts with near-zero F1; in contrast, GRACE has prior knoweldge from training on the edit set. As the sequence of edits grows, Larimar surpasses GRACE's generalization performance at around 600 edits.
%As Larimar has no knowledge of upcoming edits, it starts with lower F1, compared to GRACE that is trained on the edit set. As the sequence of edit grows, Larimar catches up at around 500 edits and achieves superior generalization to GRACE with over 1,000 sequential writes, %approximately one per fact, 
%provided that the memory capacity $K$ exceeds the fact count.  

In these experiments, we use $K=1000$, setting the memory size proportional to the number of facts to be written. %, which allows for one memory slot per independent fact. 
We also checked an alternative method (Appendix \ref{app:details}) for computing the reading and writing weights, which uses a Gaussian convolution to 
store each encoding $\mathbf{z}$ in memory location(s) corresponding to the most similar content in a reference memory $\mathbf{M}^{(\rm ref)}$, % or read from locations similar to the query Z
%assign higher weight to memory slots $k$ where the content in the $k$'th row of $\mathbf{M}^{(\rm ref)}$ is closer to the encoding $\mathbf{z}$ being stored, 
and which we found to perform better than the pseudoinverse method of \cite{pham2021generative} when there are a relatively small number of rephrasings per fact (Figure \ref{fig:zsre_seq_gen_appendix}). %to find nearby encodings in latent space; essentially just pick the closest Z in memory

%\subsection{Batch Editing}
\iffalse
we use first $2000$ Counterfact samples
sharp change at $n_{\rm edit}=K=512$
[generalization / specificity to appendix? do we have space here?]
\begin{figure}
    \centering
    \includegraphics[width=0.5\linewidth]{figures/batch-efficacy.png}
    \caption{[NEED LEGEND] Efficacy, generalization, and specificity success vs. number of edits $n_{\rm edit}$(batch size) for Larimar 6B (with $n_{\rm edit}=10,100,512,1024,2048$) compared to baselines.}
    \label{fig:batch_efficacy}
\end{figure}
\fi

% \subsection{Selective Forgetting}
% \begin{figure}[!ht]
%     \centering
%     \includegraphics[width=0.6\linewidth]{figures/larimarbatcheidt.png}
%     \caption{Batch editing accuracy on counterfact dataset. Baseline performances are taken from \cite{meng2023massediting}. Green: MEMIT, Orange: ROME, Magenta: MEND, Black: Larimar-6B.
%     }
%     \label{fig:batch}
% \end{figure}

\subsection{Selective Forgetting}
\begin{figure}[!ht]
    \centering
    \includegraphics[width=0.95\linewidth]{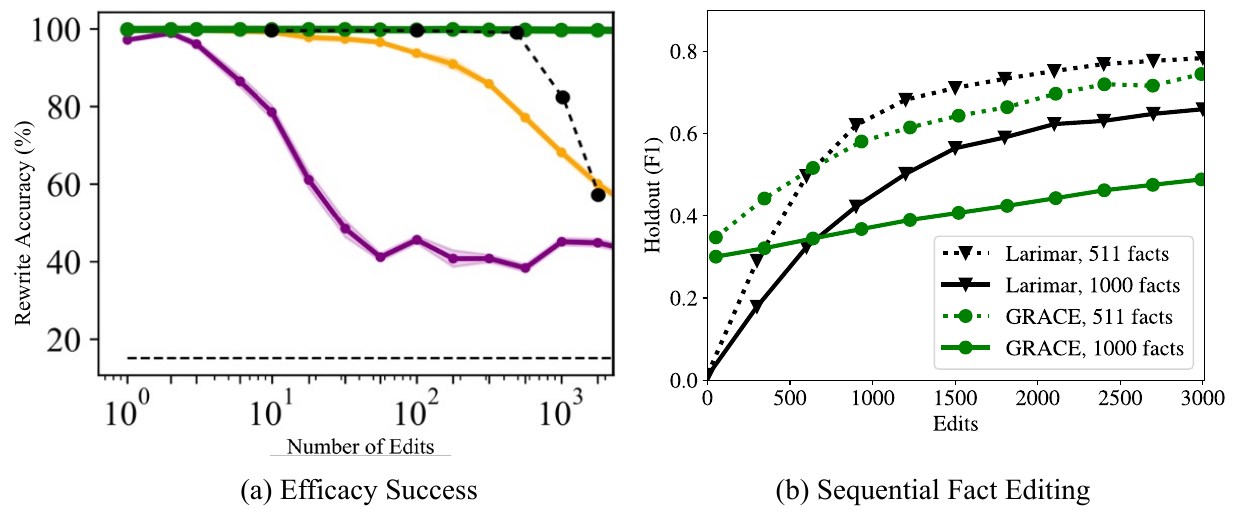}
    \vspace{-4mm}
    \caption{(a) Batch editing accuracy on Counterfact dataset. Baseline performances are taken from \cite{meng2023massediting}. Green: MEMIT, Orange: ROME, Magenta: MEND, Black: Larimar-6B. (b) Mean F1 score on a held-out set of unseen rephrasings from ZsRE  over a sequence of 3000  edits, showing Larimar's generalizes better over GRACE on two datasets with $1000$ or $511$ independent facts ($10$ and $\approx20$ rephrasings per fact, respectively).
    }
    \label{fig:batch}
\end{figure}

The goal of this section is to check if a specific fact can be selectively erased from a batch of $N$ facts that are  written to Larimar's memory in one-shot. We first checked the batch editing performance of Larimar. %if many edits can be written at once to memory and accurately retrieve from it. 
Figure \ref{fig:batch} (a) shows that the rewrite accuracy is near 100\% for up to 512 edits (eqv. to the memory size $K$) and then drops to 82\% for 1024 edits. This result shows Larimar's ability to effectively compress more than $K$ facts into its size-$K$ memory (Figure \ref{fig:batch_appendix} in appendix). This performance level is higher  when compared to baselines like MEND and ROME, but subpar compared to MEMIT \cite{meng2023massediting}, which can accurately handle a very large batch of edits at a cost of reduced editing speed (see Table \ref{tab:time}) and is also not meant to handle sequential editing. Note that Larimar's recall matches MEMIT for $N<K$ facts, and $K$ can be chosen as needed during inference (Figure \ref{fig:memory_size} in appendix). % reports the edit (rewrite) success for different values of  $K$ chosen during inference for batch size.

To test the ability of Larimar for selectively forgetting specified facts during inference, we first write $N$ facts to memory ($\alpha_i=1$ in Eq. \eqref{eq:mem_update_seq}), and then forget one fact ($\alpha_i=-1$), and also write to memory in its place (with $\alpha_i=1)$ the same fact with the answer replaced with the string ``unknown.'' We compare recall for the forgotten fact before and after the forgetting operation. To demonstrate that forgetting does not compromise other memories, we also report  the recall on the remaining $N-1$  facts in memory. % (after the forgetting operation). 
%%%When updating and reading from the memory, we use the settings described in Section~\ref{sec:seq_edit}. 
The samples used are from  the ZsRE validation set and from the Counterfact test set.
%key(encoding(query))
%value=encoding(query+completion)
%accuracy=delta(GT in response) [or, F1?]
Table \ref{forgetting_counterfact} reports these results, comparing to a $k$-shot in-context learning (see Appendix) baseline with Llama2-13B, and showing that Larimar can selectively forget using the memory updating mechanism, while retaining the remaining knowledge, whereas in-context learning struggles.

\begin{table}[!ht]
\centering
\begin{adjustbox}{width=0.49\textwidth}    
\begin{tabular}{c c c c c }
\hline
 & \multicolumn{2}{c}{\textbf{Counterfact}} & \multicolumn{2}{c}{\textbf{ZsRE}} \\
\textbf{Model} & Forgotten & Retained & Forgotten & Retained \\
\hline
Llama2 13B, $N=20$, 6-shot & 0.75 & 0.77 & 0.68 & 0.73 \\ \hline
Larimar 1.3B, $N=1$ & \textbf{0.0} & -- & \textbf{0.0} & -- \\
Larimar 1.3B, $N=K$ & \textbf{0.001} & \textbf{0.997} & \textbf{0.02} & \textbf{0.95} \\
Larimar 1.3B, $N=2K$ & 0.02 & 0.79 & 0.03 & 0.52 \\
Larimar 6B, $N=1$ & \textbf{0.0} & -- & \textbf{0.0} & -- \\
Larimar 6B, $N=K$ & \textbf{0.0} & \textbf{0.993} & 0.03 & \textbf{0.86}  \\
Larimar 6B, $N=2K$ & 0.03 & 0.71 & 0.04 & 0.50 \\
%Larimar 6B, $N=K$ & $0.011\pm0.005$ & $0.998\pm0.001$ & &  \\
%Larimar 6B, $N=1.25K$ & $0.041\pm0.006$ & $0.996\pm0.003$ &  &  \\
%Larimar 6B, $N=2K$ & $0.234\pm0.015$ & $0.906\pm0.006$ &  &  \\
%(F1) Larimar 6B, $N=2$ & $0.004\pm0.002$ & $0.992\pm0.004$ & $0.001\pm0.001$ & $0.90\pm0.01$ \\
%(F1) Larimar 6B, $N=K$ & $0.005\pm0.002$ & $0.990\pm0.001$ & $0.02\pm0.01$ & $0.90\pm0.01$ \\
%(F1) Larimar 6B, $N=1.25K$ & $0.025\pm0.003$ & $0.983\pm0.005$ & $0.031\pm0.001$ & $0.86\pm0.02$ \\
%(F1) Larimar 6B, $N=2K$ & $0.22\pm0.01$ & $0.890\pm0.005$ & $0.04\pm0.01$ & $0.58\pm0.01$ \\
%%%% Larimar 1.3B, $N=K$ & 0.01 & 0.998 & 0.01 & 0.95 \\
%%%% Larimar 1.3B, $N=1.25K$ & 0.02 & 0.99 & 0.03 & 0.89 \\
%%%% Larimar 1.3B, $N=2K$ & 0.13 & 0.90 & 0.01 & 0.48 \\

%Llama2-chat 13B & 0.58 & 0.66 & 0.46 & 0.52 \\
%%%% Llama2 70B, $N=20$, 6-shot & $0.79\pm0.03$ & $0.90\pm0.02$ &  &  \\
%%%% Llama2-chat 70B, $N=20$ & $0.63\pm0.01$ & $0.71\pm0.02$ &  &  \\
\hline
\end{tabular}
\end{adjustbox}
\caption{Fraction of facts with accurate recall, for the Counterfact and ZsRE datasets, after writing $N$ facts to memory and removing one. ``Forgotten'' and ``Retained'' indicate, respectively, recall of the fact to which forgetting  was applied, and mean recall of the $N-1$ retained facts. $K=512$ in all cases. } % Uncertainty intervals indicate $\pm$ one standard deviation over 5 runs.
\label{forgetting_counterfact}
\end{table}

\begin{table}
   \vspace{-2mm}
    \centering
    \begin{adjustbox}{width=0.45\textwidth}    
    \begin{tabular}{c c c c c}
\toprule
{\textbf{}} & {\textbf{ROME (s)}} & {\textbf{MEMIT (s)}} & {\textbf{Larimar (s)}}  & {\textbf{Larimar (b)}}\\
\midrule
%MEND (hypernetwork) & Yes & No &  3.9 & \\
Attack Success (\%) & 29.0 & 49.3 & \textbf{17.6} & \textbf{21.5}\\
\bottomrule
\end{tabular}
\end{adjustbox}
\caption{Input rephrasing attack success: Larimar-6B % Writing [prompt + ''unknown"] to Larimar-6B's memory 
in-memory writing  (single fact (s) or batch (b) mode) 
%is more effective in preventing answer leakage in generations, when compared to 
\textit{vs.} GPT-J 6B editing. %decoder via direct model editing.
}
\label{table:unlearning}
\end{table}

\textcolor{black}{We also evaluate Larimar to prevent generation of specific information by writing an empty (\textit{i.e.}, censored) response for the corresponding prompt to the memory. The baselines we consider are ROME and MEMIT, which were adapted to delete information  in \cite{patil2023sensitive}. Specifically, the  decoder $d$ was updated with an empty response objective for a given string $x$ that is known to the decoder and is aimed to be deleted, such that the probability of
an ``empty'' target string $E$ is maximized, $\text{argmax}_d\mathbb{P}[E|x,d]$. A blackbox input rephrasing attack was then used;  the presence of information of interest was checked in a number of model outputs as the model was prompted with different input rephrases. For Larimar, a single input prompt followed by ``unknown" (= empty response) is  written to the memory during inference to prevent leakage of the answer in the decoded outputs.  The attack is considered successful on a input prompt if the answer is found within a fixed number of model generations obtained using prompt rephrases. About 300 samples from Counterfact known to GPT-J 6B and Larimar's decoder were used for this experiment. 
We used 5 sample responses for each of 4 paraphrases per fact (total attack budget of 20), which were generated as prescribed in \cite{patil2023sensitive}. Table \ref{table:unlearning} shows the results, suggesting that  writing to Larimar's memory is more effective %(17.6\% attack success) 
than direct model editing methods for preventing answer leakage for a single input prompt (17.6\% attack success for Larimar, vs.  29\% and 49\%  for ROME and MEMIT, respectively). Larimar can further  restrict the response for a batch of facts in one shot -- the robustness to rephrase attacks remains still higher than baselines.  }
%update the response for multiple distinct input prompts in one-shot
% DETAILS for this experiment:
%Following \cite{patil2023sensitive}, we used a subset of facts from the CounterFact dataset which were first selected as those which were known to GPT-J, and furthermore filtered to facts which Larimar-6B was able to answer correctly (in the absence of 
%1400 filtered counterfact samples from Patil et al
%starting from these, filter to 883 known to Larimar (w/o memory)
%generate paraphrases with their dipper model
%keep the facts for which their paraphrase generating script gives at least 4 paraphrases (needed for their rephrase attack) with no duplicates
%this leaves us with 307 facts
%attack with the 4 rephrases, sampling 5 responses for each
%in this case, we just read out from memory exactly the single Z that was written

\subsection{Generalization to long input context}
%each with \VC{XXX} tokens
%$\le$ \VC{XXX}
We perform fact recall with long context using data that is not present in the base decoders pretraining corpus. For this purpose, we curated facts from CNN Fast Facts~\cite{cnnfastfact} for 2021, 2022, and 2023. We divide the input text into $T$ chunks, which is in the range of Larimar's training context window, and store each of these chunks in a separate memory $M_{i}, i=1..T$. Given a query, we address and read from each of these memories. % If the answer is found, then the search is stopped. If not, 
The readouts from these memories form the basis of the successive memory, which is then queried and read from again. This process is continued until the number of readout in the final memory is similar to Larimar's input training context window. %formed with readouts from the previous level. 
The recursive search in  latent  memory space and using readouts to construct new higher-level memory  is performed to process the long context with Larimar's memory trained on a relative small episode length.  %the last memory is used %\PD{do we also use the original query in addition} 
The  retrieved $Z_r$ from the final successor memory is passed to the decoder  for predicting  response. It should be noted that memory hierarchy is  found in hippocampus and is implicated in learning \cite{collin2015memory}. 
 
Table \ref{longgen} shows Larimar's recall does not degrade much with increasing input context length, even compared to some of  most competitive baseline LLMs trained with longer training context. We also compare with Supersizing Transformer~\cite{supersizing},  a memory-augmented model, however it did not show competitive recall performance because it was not trained to perform memory-conditioned generation. Due to memory processing in the latent space, Larimar is also efficient is terms of number of KV cache token computation compared to baseline methods. Our experiments on 128 facts show that the average time required by Larimar to read from memory is 0.36s compared to 1.44s for Mistral-7b base model.

%\textcolor{cyan}{[Move this paragraph to Related Work section?:]}

\textcolor{black}{Learning to copy from the context remains an important aspect underlying transformers' impressive language modeling and other abilities \cite{devlin2018bert, raffel2020exploring, olsson2022incontext}. %Due to the high computational complexity of transformers, search for more efficient alternative architectures is an active field of research. For example, 
LLMs with non-attention based architectures, such as state space models, often underperform \cite{gu2022efficiently, gu2023mamba} %,  On the other hand, generalized State Space Models (GSSMs) (see \cite{jelassi2024repeat} for the broad definition of GSSM), which use a fixed-size latent state  and therefore use $O(1)$ memory to predict each token, are emerging as an alternative class of language model architectures \cite{gu2022efficiently, gu2023mamba}.  GSSMs show better memory and computational efficiency, but fall behind 
transformers in language modeling, which %. This poor performance of GSSMs,
is at least partly attributed to an inability to copy from the context, as well as an inability to generalize to longer contexts, when compared to transformers \cite{jelassi2024repeat}.  Those investigations have fueled research on  hybrid architectures. The results presented here suggest that combining a  hierarchical  memory model with a generative pretrained transformer, as in Larimar, could be a promising path in that direction.  The end-to-end training of the fixed-size latent memory with the decoder in  Larimar adds an explicit state to the decoder,  writing to which helps controlling the decoding, thus allowing truthful copying from  context in a generalized manner. The memory control also provides real-time knowledge editing \textcolor{black}{as well as information leakage prevention}. Attending to the memory read-out while decoding uses $O(1)$ memory to predict each token, providing memory and computational benefits. } 

\begin{table*}[!ht]
 % \vspace{-2mm}
% \fontsize{9pt}{9pt}\selectfont
% \setlength{\tabcolsep}{0.25em} % for the horizontal padding
\label{tab:office-31-results}
\centering
% {\small\renewcommand{\arraystretch}{1.3}% for the vertical padding
    \begin{adjustbox}{width=0.85\textwidth}    
\begin{tabular}{c c c c c c c}
\toprule
 Method & Train Context & $n_{fact}=64$  & $n_{fact}=96$ & $n_{fact}=128$ & $n_{fact}=256$ \\
 \midrule
    % gptj-6b & 2048  & -  & -  &  - & - \\ 
    mistral-7b (3-shot) & 8192  & \textbf{0.98} / 2655  & \textbf{0.96} / 3495  &  0.57 / 4334 &  0.42 / 7417 \\ 
    % grep -H "Result:" logs/evaluate/fact_addition/base_model_larger/*gpt*frac_0.6*
    gpt-neox-20b (3-shot) & 2048  & 0.52 / 2366  & 0.36 / 3193  &  0.33 / 4020 &  0.35 / 7231 \\ 
    llama2-13b (3-shot) & 4096  & \textbf{0.97} / 2755  & 0.66 / 3628  &  OOM &  OOM \\
    % falcon-40b  & 2048  & -  & -  &  - &  - \\ \hline
   \hline
   \hline
   Supersizing Transformer   & 2048  & 0.39 / 1462  & 0.39 / 2249 &  0.37 / 3072 &  0.37 / 6201 \\ 
    Supersizing Transformer + filtering  & 2048  & 0.72 / 1640  & 0.71 / 2375 &  0.70 / 3110 &  0.69 / 5809 \\ \hline
    % Supersizing Transformers  & -  & -  & -  &  - \\ 
    % Larimar-1.3b (ep-6)  & 384  & 0.87 / 1565  & 0.86 / 2276  &  \textbf{0.85} / 5607 &  \textbf{0.84} / 10883 \\
    % (ep-16)
    % Larimar-1.3b (without recursive)  & 384  & 0.83  / 1565  & 0.80 / 2276  & 0.78 / 2988 &  0.73 / 5607 \\
    % Larimar-1.3b (with recursive)  & 384  & {0.89} / 1565  & {0.88} / 2276  &  \textbf{0.88} / 2988 &  \textbf{0.86} / 5607  \\
    Larimar-1.3b  & 384/1024  & {0.89} / 1565  & \textbf{0.88} / 2276  &  \textbf{0.88} / 2988 &  \textbf{0.86} / 5607  \\
    Larimar-6b  & 384/2048  & 0.82  / 1565  & 0.81 / 2276  & \textbf{0.81} / 2988 &  \textbf{0.80} / 5607 \\ \hline
    
 \bottomrule
\end{tabular}
\end{adjustbox}
\caption{\centering Novel fact addition recall rate on FastFacts. Larimar shows good  recall performance and can extrapolate to higher context length than it was trained on. Baseline models show good recall on shorter context but recall degrades significantly for longer context.}
  \vspace{-2mm}
 \label{longgen}
\end{table*}

\section{Related work}
\paragraph{Memory-augmented NNs}
External memory augmented neural networks (MANNs) were already proposed in pre-transformer era, with the aim of better learning long-term dependencies in input data \cite{weston2014memory, graves2014neural, miller2016keyvalue} showing enhanced performance in generative tasks, language modeling,  long-term planning, and sample-efficient RL, etc.
% These frameworks have shown  enhanced performance in generative tasks, language modeling,  long-term planning, and sample-efficient RL, etc. 
MANNs add a trainable slot-based memory to a recurrent neural net. An attention-based reading mechanism is typically used to compute a weighted average of memory contents. This mechanism is estimated from training data, and thus  it remains unclear how they can generalize to new data. Alternatively, Kanerva Machine \cite{wu2018kanerva}, inspired by Kanerva's sparse distributed memory model \cite{kanerva1988sparse}, views memory as a global latent variable in a generative model and aims to learn a memory dependent data prior and learnable addresses. In this framework,   the memory update and read/write are considered as Bayesian inference, i.e., the posterior parameters are updated as new data arrives. KM and its successors \cite{wu2018learning, ramapuram2022kanerva, pham2021generative}  show that %the writing speed remains an  issue for fast updating the memory. Recently, \cite{pham2021generative}  proposed to cast the memory write as a minimization problem, which provides an one-shot update mechanism for fast writing.  
these  conditional generative memory models offer better performance on image reconstuction, denoising, and generation tasks compared to variational autoencoders \cite{kingma2013auto} and memory networks \cite{bornschein2017variational}. However, to our knowledge this is the first report on  investigating how those models can adapt to LLM and aid in their knowledge updating. 

Transformers struggle with accessing and updating long-term memory \cite{fan2021addressing}. Efforts to extend input context length  struggle integrating inherent model knowledge with external facts, thereby lacking robustness \cite{li2022large, liu2023lost}. Augmenting transformers with external, non-differentiable memory and k-nearest neighbor (kNN) attention has shown promise in improving language modeling by utilizing additional context \cite{grave2017unbounded,khandelwal2019generalization}. However, kNN-augmented models face challenges in controlling memory during decoding, leading to difficulties in updating facts due to conflicts between encoded knowledge and real-time information \cite{liu2023lost, zhu2020modifying}.

\paragraph{Model Editing}
%Model editing seeks to make updates to the behavior of a pre-trained base model while satisfying three key desiderata: (i) \emph{locality}, ensuring that the model performance does not degrade elsewhere that is unrelated to edits;  (ii) \emph{generality}, ensuring that the model behaves similarly when presented with an edit reformulation; (iii) \emph{speed/scalabilty}, which is important in both sequential and batch editing situations. 
For comprehensive surveys of editing approaches see \cite{yao2023editing, zhang2024comprehensive, wang2023knowledge}.   
 Editing methods can be broadly categorized  into three categories:  `Recognition Phase', `Association Phase' and `Mastery Phase' ~\cite{zhang2024comprehensive}. %This classification essentially accounts for whether or not the  base model's parameters are affected. %~\footnote{Note that  an alternative classification is that of ~\cite{yao2023editing} : 'Cache-based', 'Meta-learning' and 'Locate and Edit'.}.  
 The `recognition phase'-targeting methods consider demonstrating right context to help the LLM output correct facts, either via  in-context demonstrations of similar examples %from training set %retrieved from a knowledge base 
 \cite{zheng2023edit}, or training an external model on edits \cite{mitchell2022memory}.% which require access to priviledged information. %  Alternatively, SERAC \cite{mitchell2022memory}  stores  edits in an external cache. Queries/inputs are classified as within/outside scope of the cache. If within, a counterfactual model produces the output; if outside, the input is routed to the base LLM. %These methods either need multiple demonstrations retrieved from a knowledge base or edit-specific training of scope classifier and counterfactual model.
The `association phase' -related editing methods consider merging new knowledge to that of the base LLM, either by patching (adding and training) error-specific neurons \cite{huang2023transformerpatcher}, or by adding  a  an adaptor storing  edit key-value pairs  to a specific LLM layer %via gradient-based update % where the values are updated by using the  finetuning loss on the model’s prediction given the edit 
\cite{hartvigsen2022aging}. %Note that both methods aim to handle a difficult but more realistic scenario than single fact editing - where  multiple edits are performed in a sequential manner and older edits are not forgotten.
  The `mastery phase' methods learn to update  base LLM's own parameters. Examples are regularized finetuning \cite{zhu2020modifying} and hypernetwork-based methods \cite{mitchell2021fast, de2021editing}. % that often decompose weight updates into low-rank components. 
  Recent works also explore the `locate-then-edit' approach:  \cite{meng2022locating, meng2022mass} first perform a causal tracing to detect which part of hidden states can be attributable to the fact and then do a rank-one update of the corresponding weight parameters to directly write in the updated fact. 

% While current model editing methods have demonstrated significant  promise \cite{yao2023editing}, they do come with certain shortcomings. They need expensive training or updating model parameters and assume   privileged  access to edits for that. Naturally, these methods fall short to systematic generalization to edits drawn from unseen  distributions. Constantly editing the LLM is not feasible, as  the edits are time- and memory-extensive \cite{mitchell2022memory}. 
%  Further,  \cite{hase2023does}  reported that fact-specific knowledge within LLM is not as localized as it is believed to be, therefore local parameter update may not be sufficient to fully enforce the  edit.  
%  When multiple edits are needed to be made  
%   either in a sequential or in a batch manner, several recent works report significant performance degradation of the model \cite{mitchell2022memory,meng2023massediting,gupta2024model, li2023unveiling, gu2024model}, revealing  forgetting of the older edits, causing knowledge distortion,  enhancing inherent knowledge inconsistencies, or loss of general abilities in LLMs. External cache/memory/codebook-based editing methods have been proposed to tackle those challenges, where base model is not directly edited. Finally, selective knowledge forgetting remains challenging for current model editing methods, as suggested in \cite{ishibashi2023knowledge}.

 Current model editing approaches, while promising \cite{yao2023editing}, face significant limitations, such as high training costs and difficulties in generalizing to new data. These methods often cannot efficiently update LLMs due to extensive time and memory requirements \cite{mitchell2022memory}. Furthermore, the assumption that knowledge within LLMs is localized has been challenged \cite{hase2023does}, indicating that simple parameter updates may not be effective for comprehensive edits. The performance of LLMs degrades with multiple edits, leading to issues like knowledge forgetting and distortion \cite{mitchell2022memory,meng2023massediting,gupta2024model, li2023unveiling, gu2024model}. Alternatives like external cache or memory-based editing have been proposed to circumvent direct model modifications, yet challenges in selectively forgetting outdated or sensitive knowledge persist \cite{ishibashi2023knowledge, patil2023sensitive}.
 
% Different from the above-mentioned works, here  we showcase for the first time the generative memory model augmentation to a LLM. Further, we demonstrate that the dynamic updating of the generative memory  can enable fact editing and forgetting, as well as  input length generalization of a LLM decoder in a training-free and accurate manner.
% Our work is distinct  from existing LLM editing methods that adopt the classical view of generative LLM or the linear layer within as an associative memory \cite{meng2022locating} and therefore  directly update some of LLM parameters to reflect new knowledge \cite{ meng2022locating, meng2022mass}. 
% Our work is also different from  model editing methods that update a slot-based external memory 
% \cite{han-etal-2023-improving, hartvigsen2022aging}. Further, Larimar does not need multiple demonstrations that are similar to the desired edit in order to control the decoder as in \cite{zheng2023edit}.

Different from the above-mentioned works, we present a novel approach to augment LLMs with generative memory, enabling dynamic editing and adaptation without retraining. This differs from existing works that update LLM parameters \cite{ meng2022locating, meng2022mass} or external memories \cite{han-etal-2023-improving, hartvigsen2022aging}, or requires multiple in-context demonstrations \cite{zheng2023edit}.

Larimar's forgetting operation does not use negative examples to fine-tune LLMs for unlearning \cite{yu2023unlearning}. Neither Larimar requires tailored fine-tuning \cite{eldan2023whos} or inserting extra layers \cite{chen2023unlearn}, and is complimentary to in-context unlearning (e.g., \cite{pawelczyk2023context}) for fact forgetting.
 %Instead,  Larimar's memory undergoes a fast training-free update as new edits arrive, in contrast to existing editing methods. Finally, the inference-time forgetting operation in Larimar is unique. %shares significant dissimilarities with that line of work, as (i) it's external memory is associative in nature: a set of $n$ facts is written in and read out from a fixed size memory,  
%Further, read out from Larimar's memory conditions the decoding to enable outputting the edit, rather than directly changing LLM parameters via training or rank-one updating. 
 % As new data (different from training) arrives during inference,  encoding new data in memories and decoding data from memories are postulated as Bayesian updates for which an equivalent minimization problem is proposed. This minimization problem, which corresponds to solving a linear system of equations, is efficiently done via computing matrix pseudo inverses. This Bayesian updating mechanism enables Larimar to handle unseen edits in a training-free manner, which is not the case for existing editing methods.

  %Previous works on generative memory models %, and more generally, associative memory models,
  %have tested model's performance mostly on image reconstruction, denoising, and generation tasks. To our knowledge, this is the first work that proposes that trains a hierarchical conditional generative memory model for language and tackle LLM editing in a more bio-inspired manner via dynamically updating the generative distributed memory.   

  %\subsection{rag}

\section{Conclusions}
\label{sec:conclusion}
 In  this work, we propose augmenting LLMs with a dynamically updatable and distributed episodic memory as a means to online knowledge adaptation. By exploiting a one-shot memory update mechanism, combined with memory-conditioned decoding, the proposed framework shows \emph{accurate},  \emph{precise}, \emph{robust}, and \emph{significantly faster} editing performance compared to baselines in single-fact, as well as the challenging sequential %and batch 
 editing experiments. % when evaluated on existing fact editing benchmarks.  
 We exploit the same memory updating mechanism to  enable a \emph{fast and selective} fact forgetting operation, \textcolor{black}{as well as an effective information deletion mechanism}.  We also provide a simple approach for handling long input context by recursively reading from Larimar's memory space, revealing better fact recall from long input context by Larimar when compared to state-of-the-art LLMs trained with a much larger training context window.  The proposed framework thus provides a simple, general, and principled approach to update LLMs in real-time by coupling them with an  adaptable episodic memory control. %mechanism extend this architecture for   that is dynamically updatable, emonstrate effectiveness of a global episodic memory in model-agnostic, fact tracing and training- free, online editing. general and simple approach. 

%The above statement can be used verbatim in such cases, but we encourage authors to think about whether there is content which does warrant further discussion, as this statement will be apparent if the paper is later flagged for ethics review.

\textcolor{black}{
One obvious limitation of current Larimar architecture is being able to only handle shorter-lengh facts. In addition, the training is currently limited to sentence completion tasks. Future works will include expanding Larimar to modeling longer sentences and to more tasks like question answering, and summarization. Further, we will subject Larimar to more challenging tasks during inference, whereas the knowledge update in latent memory is needed for the model to navigate the task, for example in conversational setting.}

\clearpage

\section*{Impact Statement} 
This paper presents work whose goal is to advance the field of machine learning and large language models. There are many potential societal consequences of our work, none which we feel must be specifically highlighted here.
\bibliography{ref} %Prints bibliography
\clearpage

%------------------------------------
%   APPENDIX
%------------------------------------
\appendix

%\subsection{Fast memory write and no retraining enable efficient knowledge update with Larimar}

%\todo{\paragraph{Dataset and preprocessing} ...}

%\todo{\paragraph{Experiment setup} Parameters, number of iteratione etc}

%\begin{figure}[!h]
%    \centering
%    \includegraphics[width=\linewidth]{figures/edit-comp.png}
%    \caption{}
%    \label{fig:edit-comp}
%\end{figure}

%\begin{figure}[!h]
%    \centering
%    \includegraphics[width=0.5\linewidth]{figures/edit-wallclock.png}
%    \caption{Wall clock time for each edit method for performing 10 edits
%}
%    \label{fig:edit-wallclock}
%\end{figure}
\section{Baselines}
\paragraph{FT}
Fine-Tuning (FT) uses Adam optimization with early stopping, focusing on adjusting $mlp_{proj}$ weights in one layer to optimize the training loss. 

\paragraph{FT+L}
Constrained fine-tuning (FT+L), as in \cite{zhu2020modifying}, authors apply an $L_\infty$ norm constraint  by clamping weights no to exceed $\epsilon$ range at each gradient step. They chose layer 0 and $\epsilon = 5\times 10^{-4}$ for GPT-2, and $\epsilon=5\times 10^{-5}$ for GPT-J. 

\paragraph{KN}
This is a method by \cite{dai2021knowledge} which selects neurons that are associated with knowledge expression via gradient-based attributions, and then modifies MLP at the rows corresponding to those neurons by adding scaled embedding vectors.

\paragraph{KE}
Knowledge editor (KE) \cite{de2021editing} learn an LSTM sequence model that uses gradient information to predict rank-1 weight changes to the model. KE-CF / KE-ZsRE is additionally trained model on training set of CounterFact / ZsRE dataset. 

\paragraph{MEND}
Model Editor Networks with Gradient Decomposition (MEND) \cite{mitchell2021fast} learn a rank-1 decomposition of the negative log likelihood gradient with respect to some subset of parameters . Similarly, MEND-CF / MEND-ZsRE is additionally trained model on training set of CounterFact / ZsRE dataset.

\paragraph{ROME}
Rank-One Model Editing (ROME), proposed by \cite{meng2022locating}, treats MLP module as a key-value store. To add a new key-value pair, ROME applies a rank-one modification to the weights of the MLP, adding the new information directly. 

\paragraph{IKE}
In-context Knowledge Editing (IKE) \cite{zheng2023edit} defines three types of demonstration formatting templates including copy, update, and retain, which guide model to edit knowledge facts by in-context learning (ICL). The parameters of the model are not updated. 

\paragraph{PROMPT}
Similar to IKE \cite{zheng2023edit} but simply prepends new fact to the LLM prompt. The parameters of the model are also not updated. 

%\paragraph{PMET}
%Precise Model Editing in a Transformer (PMET), proposed by \cite{li2023pmet}, simultaneously optimizes transformer component (multi-head self-attention and
%feed-forward layers (FFN)) hidden states, while only using the optimized hidden states of FFN to precisely update FFN weights. 

\paragraph{MEMIT} 
MEMIT aims direct model editing via fact tracing and followed by parameter editing. It is an expanded version of  ROME, which   enables the editing of large amounts of factual data through the updating of a sequence of MLP layers.

\paragraph{SERAC}
SERAC is a retrieval-based editing algorithm which uses a retrieval-based component consisting of an external memory that contains an explicit cache of edits. In addition, an edit scope classifier and a counterfactual model are trained using these edits. If the new input is identified as within the scope, the output from the counterfactual model is returned. If not, the base model is used for generation.  

\paragraph{ICL}
To compare to In-Context Learning (ICL) as a baseline method in Table \ref{forgetting_counterfact}, we use a prompt which consists of $N$ facts, half of which are marked with a prefix string (e.g. ``[UNKNOWN]''), followed by $K$ examples of questions and answers (prior to a final query to the model), half of which correspond to facts marked with the prefix string, which replaces the answer, indicating that the fact should be treated as forgotten.

In Table \ref{tab:compare_methods}, we report the wall clock time for a single edit (averaged over 10 edits) on the CounterFact dataset for ROME \cite{meng2022locating} and GRACE \cite{hartvigsen2022aging} that were computed using the EasyEdit~\cite{wang2023easyedit} framework with a single A100 (80G) GPU. Below we provide wall clock time of different existing editing methods, as reported by 
\cite{yao2023editing}.

\begin{table}[!h]
\begin{tabular}{lc}
\hline Editor & COUNTERFACT \\
\hline FT & $35.94 \mathrm{~s}$ \\
SERAC & $5.31 \mathrm{~s}$ \\
CaliNet & $1.88 \mathrm{~s}$ \\
T-Patcher & $1864.74 \mathrm{~s}$ \\
KE & $2.20 \mathrm{~s}$ \\
MEND & $0.51 \mathrm{~s}$ \\
KN & $225.43 \mathrm{~s}$ \\
ROME & $147.2 \mathrm{~s}$ \\
MEMIT & $143.2 \mathrm{~s}$ \\
%GRACE & $1.3 \mathrm{~s}$\\
%Larimar-1.3B & $\mathbf{0 . 1 5 6} \mathrm{s}$ \\
%Larimar-6B (A100) & $\mathbf{1.88} \mathrm{s}$ \\
\hline
\end{tabular}
\caption{Wall clock time for each edit method for performing 10 edits from CounterFact benchmark, as reported in \cite{yao2023editing}. }%**"Wall clock time for each edit method conduct- ing 10 edits on GPT-J using one 2×V100 (32G). The calculation of this time involves measuring the duration from providing the edited case to obtaining the post- edited model". from https://aclanthology.org/2023.emnlp-main.632.pdf. Grace performance is on GPT2-xl+wikibio on 1 48GB NVIDIA RTX A6000 GPU. Our performance is on 80 gb a100. this paper gives some baselines (gpt-2xl) for 1000 sequential edits zsre? on a100 https://aclanthology.org/2023.findings-emnlp.749.pdf}
\label{tab:time}
\end{table}

\subsection{Prompt Example For Forgetting with ICL experiments}
Below, we show an example ICL prompt for $N=20$ and $K=6$, used for Llama2-13B in Table \ref{forgetting_counterfact}:
\newline
\newline
Facts: \newline
Gaston Palewski writes in French \newline
The domain of work of Hermann Klaatsch is anatomy \newline
[UNKNOWN] 2 Minute Drill is to debut on ESPN \newline
[UNKNOWN] Immanuel Wallerstein works in the area of sociology \newline
[UNKNOWN] Nissan Skyline, created by Nissan \newline
Feng Fei-fei, a citizen of Taiwan \newline
Michie Mee works as actor \newline
[UNKNOWN] Antonio Palomino's life ended in Madrid \newline
[UNKNOWN] Film Forum is within Manhattan \newline
[UNKNOWN] Phoenix Sky Harbor International Airport is located in Phoenix \newline
Abraham \& Straus formed in Brooklyn \newline
The domain of activity of Joseph Schumpeter is economics \newline
George Buza's profession is an actor \newline
[UNKNOWN] Velayudham originated in India \newline
Sophie Calle, a native French \newline
In Nunavut, the language spoken is French \newline
[UNKNOWN] The Wide Country is to debut on NBC \newline
[UNKNOWN] The mother tongue of Porfiry Ivanov is Russian \newline
In Kiiminki, they understand Finnish \newline
[UNKNOWN] Pachomius the Great succumbed at Egypt \newline
 \newline
Input: The domain of work of Hermann Klaatsch is \newline
Output: anatomy \newline
Input: 2 Minute Drill is to debut on \newline
Output: UNKNOWN \newline
Input: Immanuel Wallerstein works in the area of \newline
Output: UNKNOWN \newline
Input: Nissan Skyline, created by \newline
Output: UNKNOWN \newline
Input: Feng Fei-fei, a citizen of \newline
Output: Taiwan \newline
Input: Michie Mee works as \newline
Output: actor \newline
Input: Gaston Palewski writes in \newline
Output: \newline

\section{Ablation experiments on CounterFact single fact editing}

In Table \ref{tab:counter_ablate_para} we show that when Larimar has access to additional fact paraphrases, its paraphrase performance increases from 88.4 to 93.6. Note that in this setup the average number of added paraphrased facts is one or two and we queried the model with paraphrased prompts unseen by the memory. Also, observe that the use of the scope detector for query detection is crucial for the model's performance to properly handle the neighborhood prompts.
\iffalse
\begin{table}[!ht]
    \centering
    \begin{adjustbox}{width=0.48\textwidth}    
    \begin{tabular}{l r r r}
        \toprule
        \textbf{Editor} & \textbf{Edit Success} & \textbf{Paraphrase}  & \textbf{Neighborhood} \\
        \midrule
\textbf{Larimar-6B w/ scope} & 99.6 & 76.5 & 80.2 \\
\textbf{Larimar-6B +para} & 99.6 & 82.8 & 80.6 \\
\textbf{Larimar-6B +para, no scope} & 99.6 & 88.7 & 16.3 \\\bottomrule
    \end{tabular}
\end{adjustbox}
\caption{Single fact edit evaluation on CounterFact dataset. Larimar-6B base is the baseline which includes only a single fact in the memory and uses in-scope query detector. Larimar-6B +para is the version which adds into the memory on average one additional paraphrased fact.
}
\label{tab:counter_ablate_para}
\end{table}
\fi
\begin{table}[!ht]
    \centering
    \begin{adjustbox}{width=0.48\textwidth}    
    \begin{tabular}{l r r r}
        \toprule
        \textbf{Editor} & \textbf{Edit Success} & \textbf{Paraphrase}  & \textbf{Neighborhood} \\
        \midrule
\textbf{Larimar-6B w/ scope} & 99.6 & 88.4 & 80.4 \\
\textbf{Larimar-6B w/ scope + 1 rephrase} & 99.7 & 92.9 & 79.3 \\
\textbf{Larimar-6B w/ scope + 2 rephrases} & 99.8 & 93.6 & 79.2 \\
\textbf{Larimar-6B w/o Scope} & 99.6 & 93.6 & 13.7 \\
\textbf{Larimar-6B w/o Scope + 1 rephrase} & 99.7 & 95.9 & 11.0 \\
\textbf{Larimar-6B w/o Scope + 2 rephrases} & 99.8 & 96.2 & 10.5 \\\bottomrule
    \end{tabular}
\end{adjustbox}
\caption{Single fact edit valuation on CounterFact dataset. Larimar-6B w/ Scope  is the baseline which includes only a single fact in the memory and uses in-scope query detector. Larimar-6B + rephrase is the version which adds into the memory on average one  or more additional paraphrased facts during test and queries the memory with an unseen rephrased prompt. Results without  a scope detector (w/o Scope) are also reported.}.
\label{tab:counter_ablate_para}
\end{table}

In Table \ref{tab:ablation_gpt2xl} and \ref{tab:ablation_gptj} we provide ablation results on Larimar by varying different learning parameters and architectural components of the model and observing performance on CounterFact dataset. In Table \ref{tab:ablation_gpt2xl} the ablation results for GPT-2 Large  based model are presented. Here we examined three different training configurations: 
\begin{itemize}
    \item C1: Episode length 6, observation noise 0.0001, trained for 2 epochs
    \item C2: Episode length 20, observation noise 0.000001, trained for 4 epochs
    \item C3: Episode length 16, observation noise 0.000001, trained for 2 epochs    
\end{itemize}
Note that the model reported in Table \ref{tab:zsre} in main paper is based on configuration C3. Moreover, we looked at three versions of the Larimar architecture: Original Larimar, Larimar without Scope detector and Larimar without memory. As can be seen, configuration C3 had some edge in performance. The effect of removing scope detector is reflected in drop of the neighborhood score. This is expected since now the model reroutes the prompts from the unconstrained decoder to the memory-constrained one, where the memory influence makes it harder to cover prompts unrelated to in-memory content. On the other hand, removing memory module results in significant decrease in edit success and paraphrasing, as now the model has no knowledge about introduced knowledge facts, at the same time its general language abilities are intact and performing well as reflected in high neighborhood score.

\begin{table}
\centering
\begin{adjustbox}{width=0.4\textwidth}
\begin{tabular}{clrrrrrr}
\toprule
\multirow{3}{*}{\textbf{Config}} & \multirow{3}{*}{\textbf{Editor}} & \multicolumn{6}{c}{\textbf{Metrics}} \\
\cmidrule(lr){3-8}
& & \multicolumn{2}{c}{\textbf{Edit Success}} & \multicolumn{2}{c}{\textbf{Paraphrase}} & \multicolumn{2}{c}{\textbf{Neighb}} \\
\cmidrule(lr){3-4}\cmidrule(lr){5-6}\cmidrule(lr){7-8} 
& & S & M & S & M & S & M \\
\midrule
& Larimar & 100.0 & 99.7 & 81.3 & 48.9 & 75.5 & 2.1 \\
C1 & No Scope & 100.0 & 99.8 & 81.9 & 48.66 & 28.5 & -27.4 \\ 
& No Memory & 23.3 & -4.4 & 26.5 & -3.5 & 77.7 & 4.7 \\ 
\midrule
& Larimar & 100.0 & 99.9 & 80.3 & 49.1 & 74.7 & 1.9 \\
C2 & No Scope & 100.0 & 99.9 & 80.3 & 51.2 & 24.5 & -36.9 \\ 
& No Memory & 20.6 & -4.9 & 24.5 & -4.1 & 78.9 & 5.4 \\ 
\midrule
& Larimar & 100.0 & 99.8 & 85.4 & 56.7 & 74.7 & 1.6 \\
C3 & No Scope & 100.0 & 99.9 & 87.7 & 57.2 & 15.1 & -46.1 \\ 
& No Memory & 21.6 & -4.8 & 25.4 & -3.8 & 78.4 & 5.0 \\ 
\bottomrule
\end{tabular}
\end{adjustbox}
\caption{Ablation results for Larimar-1.3B using CounterFact dataset}
\label{tab:ablation_gpt2xl}
\end{table}

In Table \ref{tab:counterfact_abl} we show ablation results with different scope detectors (detects whether a given prompt is related or not to the facts written in memory). ESD (externally trained scope detector) shows overall good performance across all three metrics (edit success, paraphrase and neighborhood). ISD (internal trained scope detector on train st of CounterFact dataset) shows some improvement in edit success and paraphrase while dropping the performance on neighborhood metric, which is expected as now it is less accurate in classifying the neighborhood prompts as out of scope.  

In Table \ref{tab:ablation_gptj} the ablation results for GPT-J based model represent results for the following five training configurations:
\begin{itemize}
    \item C1: Episode length 5, no KL loss, trained for 5 epochs
    \item C2: Episode length 16, noise level 1e-4, trained for 8 epochs    
    \item C3: Episode length 16, noise level 1e-4, no KL loss, trained for 8 epochs        
    \item C4: Episode length 8, noise level 1e-4, trained for 8 epochs         
    \item C5: Episode length 8, noise level 1e-4, no KL loss, trained for 8 epochs          
\end{itemize}
Note that the model reported in Table \ref{tab:counterfact} in main paper is based on configuration C1.
Similarly as before, we looked at architectural changes which included the removal of scope detector and memory block. We observed that configuration C2 performed the worst, while C1 had overall better performance. Moreover, the experiments again confirmed the benefit of scope detector and the effect of memory unit.

% \begin{table}
% \centering
% \begin{adjustbox}{width=0.4\textwidth}
% \begin{tabular}{clrrrrrr}
% \toprule
% \multirow{3}{*}{\textbf{Config}} & \multirow{3}{*}{\textbf{Editor}} & \multicolumn{6}{c}{\textbf{Metrics}} \\
% \cmidrule(lr){3-8}
% & & \multicolumn{2}{c}{\textbf{Edit Success}} & \multicolumn{2}{c}{\textbf{Paraphrase}} & \multicolumn{2}{c}{\textbf{Neighb}} \\
% \cmidrule(lr){3-4}\cmidrule(lr){5-6}\cmidrule(lr){7-8} 
% & & S & M & S & M & S & M \\
% \midrule
% & Larimar & 99.6 & 96.0 & 76.3 & 22.1 & 80.2 & 3.9 \\
% C1 & No Scope & 99.6 & 96.1 & 83.6 & 23.6 & 10.4 & -32.8 \\ 
% & No Memory & 15.8 & -6.8 & 18.6 & -6.8 & 83.6 & 6.9 \\ 
% \midrule
% & Larimar & 42.4 & 3.4 & 37.8 & -2.7 & 82.9 & 6.9 \\ 
% C2 & No Scope & 42.5 & 3.4 & 38.8 & -2.7 & 67.8 & 5.7 \\ 
% & No Memory & 15.2 & -7.0 & 18.3 & -6.3 & 83.2 & 6.9 \\  
% \midrule
% & Larimar & 99.9 & 98.9 & 68.0 & 10.8 & 79.9 & 3.1 \\ 
% C3 & No Scope & 99.9 & 99.0 & 81.3 & 11.9 & 7.2 & -48.1 \\ 
% & No Memory & 15.0 & -6.6 & 18.5 & -6.2 & 83.6 & 6.5 \\ 
% \midrule
% & Larimar & 91.1 & 70.8 & 66.1 & 16.1 & 81.6 & 5.9 \\
% C4 & No Scope & 91.0 & 70.8 & 67.9 & 16.3 & 27.8 & -5.4 \\ 
% & No Memory & 15.4 & -6.9 & 18.1 & -6.0 & 83.2 & 6.6 \\ 
% \midrule
% & Larimar & 99.9 & 98.9 & 72.2 & 15.1 & 79.7 & 3.3 \\
% C5 & No Scope & 99.9 & 99.0 & 82.9 & 16.5 & 6.6 & -50.4 \\ 
% & No Memory & 14.3 & -6.9 & 18.8 & -6.3 & 83.5 & 6.8 \\ 
% \bottomrule
% \end{tabular}
% \end{adjustbox}
% \caption{Ablation results for Larimar-6B using CounterFact dataset}
% \label{tab:ablation_gptj}
% \end{table}

\begin{table}
\centering
\begin{adjustbox}{width=0.4\textwidth}
\begin{tabular}{clrrrrrr}
\toprule
\multirow{3}{*}{\textbf{Config}} & \multirow{3}{*}{\textbf{Editor}} & \multicolumn{6}{c}{\textbf{Metrics}} \\
\cmidrule(lr){3-8}
& & \multicolumn{2}{c}{\textbf{Edit Success}} & \multicolumn{2}{c}{\textbf{Paraphrase}} & \multicolumn{2}{c}{\textbf{Neighb}} \\
\cmidrule(lr){3-4}\cmidrule(lr){5-6}\cmidrule(lr){7-8} 
& & S & M & S & M & S & M \\
\midrule
& Larimar & 99.6 & 96.0 & 87.6 & 53.9 & 80.6 & 4.4 \\
C1 & No Scope & 99.6 & 96.1 & 94.6 & 55.5 & 15.7 & -17.1 \\ 
& No Memory & 15.8 & -6.8 & 18.6 & -6.8 & 83.6 & 6.9 \\ 
\midrule
& Larimar & 69.8 & 11.5 & 59.2 & 5.7 & 82.7 & 6.8 \\ 
C2 & No Scope & 70.8 & 11.6 & 64.4 & 6.4 & 62.7 & 4.2 \\ 
& No Memory & 15.2 & -7.0 & 18.3 & -6.3 & 83.2 & 6.9 \\  
\midrule
& Larimar & 99.6 & 98.9 & 88.8 & 59.1 & 80.3 & 3.6 \\ 
C3 & No Scope & 99.9 & 99.0 & 95.3 & 60.6 & 15.3 & -22.1 \\ 
& No Memory & 15.0 & -6.6 & 18.5 & -6.2 & 83.6 & 6.5 \\ 
\midrule
& Larimar & 91.0 & 69.7 & 78.9 & 34.7 & 81.6 & 5.9 \\
C4 & No Scope & 91.0 & 69.7 & 83.9 & 35.4 & 29.8 & -4.2 \\ 
& No Memory & 15.4 & -6.9 & 18.1 & -6.0 & 83.2 & 6.6 \\ 
\midrule
& Larimar & 99.9 & 98.9 & 88.7 & 59.9 & 80.1 & 3.7 \\
C5 & No Scope & 99.9 & 98.9 & 94.9 & 61.4 & 15.7 & -22.8 \\ 
& No Memory & 14.3 & -6.9 & 18.8 & -6.3 & 83.5 & 6.8 \\ 
\bottomrule
\end{tabular}
\end{adjustbox}
\caption{Ablation results for Larimar-6B using CounterFact dataset}
\label{tab:ablation_gptj}
\end{table}

\begin{table}
\centering
\begin{adjustbox}{width=0.42\textwidth}
\begin{tabular}{lrrrrrr}
\toprule
\multirow{3}{*}{\textbf{Editor}} & \multicolumn{2}{c}{\textbf{Edit Success}} & \multicolumn{2}{c}{\textbf{Paraphrase}} & \multicolumn{2}{c}{\textbf{Neighborhood}} \\
\cmidrule(lr){2-3}\cmidrule(rr){4-5}\cmidrule(lr){6-7} & S & M & S & M & S & M \\
\midrule
Larimar (ESD) & 99.6 & 95.9 & 87.6 & 25.9 & 80.6 & 4.3 \\
Larimar(ISD-ep4) & 99.6 & 96.6 & 94.6 & 55.5 & 15.7 & -17.0 \\
Larimar(ISD-ep8) & 99.6 & 94.9 & 86.6 & 46.9 & 45.7 & -5.9 \\ \bottomrule
\end{tabular}
\end{adjustbox}
\caption{Ablation experiment on Larimar-6B using CounterFact dataset with different scope detectors: external vs internal  (trained on counterfact data).}
\label{tab:counterfact_abl}
\end{table}

%Larimar (ESD) & 99.6 & 96.0 & 87.6 & 53.9 & 80.6 & 4.4 \\
%Larimar(ISD-ep8) & 99.6 & 94.4 & 80.4 & 40.6 & 28.8 & -11.8 \\ 

\section{ZsRE single fact editing experiments and ablations}

We evaluated Larimar on the ZsRE benchmark \cite{zsre2017}, a QA dataset for relation extraction through reading comprehension. See Table \ref{tab:zsre} for details.

The test-time augmentation of memory with additional paraphrases boosts generalization from 70.4\% to 82.2\%, when two rephrases written in memory. In absence of scope detector, the same approach of augmenting memory with two additional rephrases provide an additional 1-2\% increase in generalization, whereas neighborhood specificity is affected significantly, irrespective of the dataset.

%Larimar demonstrates effective editing and paraphrasing on ZsRE, with comparable or slightly lower performance in the Neighborhood category, maintaining consistent results across GPT-2 and GPT-J decoders, underscoring its model-agnostic editing capabilities.

\begin{table}
    \centering
    \begin{adjustbox}{width=0.48\textwidth}    
    \begin{tabular}{l r r r}
        \toprule
        \textbf{Editor} & \textbf{Edit Success} & \textbf{Paraphrase}  & \textbf{Neighborhood} \\
        \midrule
GPT-2 XL & 22.2  & 21.3  & 24.2 \\\midrule
FT & 99.6 & 82.1  & 23.2 \\
FT+L & 92.3  & 47.2 & 23.4 \\
KE & 65.5  & 61.4  & 24.9 \\
KE-zsRE & 92.4  & 90.0  & 23.8 \\
MEND & 75.9 & 65.3 & 24.1 \\
MEND-zsRE & 99.4  & 99.3  & 24.1 \\
ROME & 99.8  & 88.1  & \textbf{24.2} \\
\textbf{Larimar-1.3B} & 98.1 & 81.6 & 19.7 \\
\midrule\midrule
GPT-J & 26.4  & 25.8  & 27.0 \\
\midrule
%FT & 69.6 & 64.8 & 24.1 \\
%MEND & 99.6 & 18.6 & 22.4 \\
%ROME & 21.0 & 19.6 & 0.9 \\
ROME & 99.8 & 95.9 & 27.2 \\
%PMET & 96.9 & 90.6 & 26.7 \\\midrule
\textbf{Larimar-6B} & 94.5 & 70.4 & 25.1 \\
\textbf{Larimar-6B + 2 rephrases} & 94.5 & 82.2 & 25.1 \\\bottomrule
    \end{tabular}
\end{adjustbox}
\caption{Single fact edit valuation on ZsRE dataset. Larimar closely matches or outperforms gradient based,  locate-then-edit based, and ICL baselines with training-free memory-conditioned generation.
% Highlighted in each column are top-two best systems. %The baselines models are all based on GPT-2 XL. We compare them against Larimar 1.3B and Larimar 6B.
}
\label{tab:zsre}
\end{table}

\begin{table}
    \centering
       \begin{adjustbox}{width=0.48\textwidth}   
    \begin{tabular}{l|ccc}
    \hline
   zSRE results	& Edit Success	& Paraphrase & 	Neighb\\
   \hline
Larimar (W/ Scope) + 1 rephrase	& 91.8	& 77.9	& 25.1\\
Larimar (W/ Scope) + 2 rephrase	& 91.8	& 82.2	& 25.1\\
Larimar (W/O Scope) + 1 rephrase &	92.1 &	79.4 &	8.6\\
Larimar (W/O Scope) + 2 rephrase &	92.1 &	83.4 &	8.5\\
  \hline
    \end{tabular}
    \end{adjustbox}  
    \caption{Knowledge editing generalization results with and without additional rephrases and scope detectors on zSRE.}
    \label{tab:ke_gen_zSRE}
\end{table}

\section{Additional Counterfact Batch editing Results}

\begin{figure}
    \centering
    \includegraphics[width=0.9\linewidth]{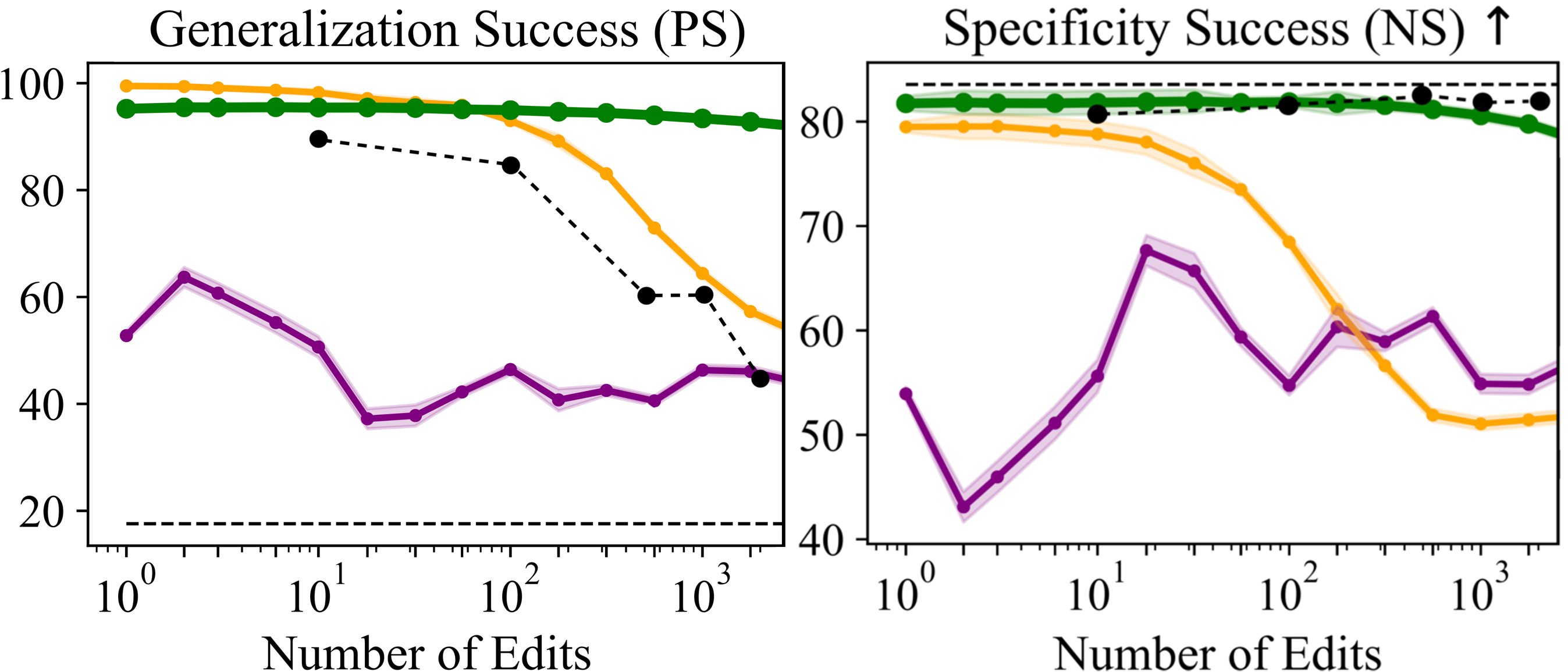}
    \caption{Batch editing on CounterFact dataset. Baseline performances are taken from \cite{meng2023massediting}. Green: MEMIT, Orange: ROME, Magenta: MEND, Black: Larimar-6B.
    }
    \label{fig:batch_appendix}
\end{figure}
\textcolor{black}{Figure \ref{fig:batch_appendix}
shows the generalization and neighborhood specificity comparison of Larimar with three baselines, MEMIT, ROME, and MEND. The result  indicates  Larimar maintains generalization performance of single fact editing up to a batch size of 512, for larger batches the performance drops. The neighborhood specificity of Larimar, thanks to the  use of the scope detector, remains very high for all batch sizes. }

\begin{figure}
    \centering
    \includegraphics[width=0.9\linewidth]{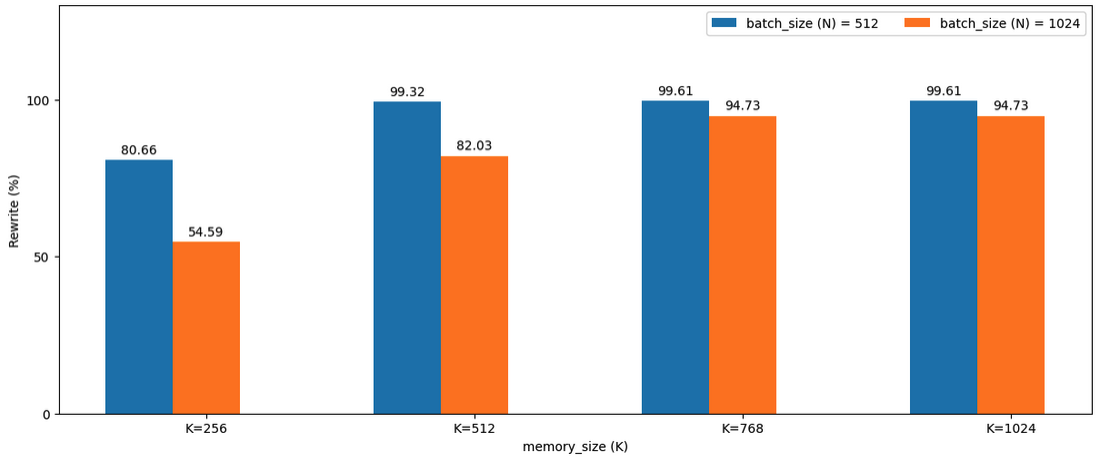}
    \caption{Batch editing on CounterFact dataset with different memory slot size $K$.
    }
    \label{fig:memory_size}
\end{figure}
We use a memory of size $K \times C$, where K=512 and C=768 throughout the manuscript, unless otherwise stated.  For the sequential editing experiments, where there were 1000 facts to be stored in memory, $K$ was set to 1000. Figure \ref{fig:memory_size} shows  the edit performance change as a function of the number of updates with different memory sizes. Results suggest that for $N=K=512$, rewrite accuracy is ~99\%. For $K=N>512$, the rewrite accuracy is slightly lower ~94\%, likely because Larimar was trained with K=512. For $N>K$, where K is smaller than 512, we find rewrite accuracy to be around 80\% if $N=2K$ and around 54\% if $N=4K$.
\section{Additional Experimental Details}
\label{app:details}

In several experiments, we compute both reading and writing weights using a Gaussian filter, as follows.
Given an encoding $\mathbf{z}$ to be written to memory, and reference memory matrix $\mathbf{M}^{(\rm ref)}$, we define the writing weight element $w_k$ at memory slot $k$ as
\begin{equation}\label{eq:gaussian_conv}
    %    w_k = \frac{\exp(-\alpha d(\mathbf{z},\mathbf{M}_{k,:}|\mathbf{M}))}{\sum_{j=1}^K \exp(-\alpha d(\mathbf{z},\mathbf{M}_{j,:}|\mathbf{M}))},
    w_k(\mathbf{z}|\mathbf{M}^{(\rm ref)}) \propto \exp\Big(-\frac{||\mathbf{z}-\mathbf{M}^{(\rm ref)}_{k,:}||_2^2}{2\alpha\sigma^2(\mathbf{z}|\mathbf{M}^{(\rm ref)})}\Big),
\end{equation}
where ``$\propto$'' implies that we normalize the weight vectors such that $\sum_{k=1}^K w_k=1$, 
$\alpha$ is a parameter which controls the entropy or sparsity of the weights ($\mathbf{w}$ becomes a one-hot vector, or multinomial distribution with zero entropy, as $\alpha\rightarrow0$), %we use the squared Euclidean distance as a similarity metric,
and
%\begin{equation} %d(\mathbf{z},\mathbf{z}'|\mathbf{M}):=\frac{||\mathbf{z}-\mathbf{z}'||_2^2}{2\sigma^2(\mathbf{z}|\mathbf{M})},
%\end{equation}
we choose the width function $\sigma(\mathbf{z}|\mathbf{M}^{(\rm ref)})$ to be the distance from $\mathbf{z}$ to the nearest neighbor row in $\mathbf{M}^{(\rm ref)}$,
\begin{equation}
    \sigma(\mathbf{z}|\mathbf{M}^{(\rm ref)}) := \min_k ||\mathbf{z}-\mathbf{M}^{(\rm ref)}_{k,:}||_2.
\end{equation}
Eq. \eqref{eq:gaussian_conv} assigns a lower weight $w_k$ to memory locations $k$ for which the distance $||\mathbf{z}-\mathbf{M}^{(\rm ref)}_{k,:}||_2$ is large compared to the nearest-neighbor distance $\sigma(\mathbf{z}|\mathbf{M}^{(\rm ref)})$.
%is small (large) the Gaussian width is decreased (increased).

\textbf{Sequential editing experiments.}

For the sequential editing experiments reported in Table \ref{tab:zsre_seq} and Figure \ref{fig:batch} (b), we set $K=1000$ and use a fixed reference memory $\mathbf{M}^{(\rm ref)}$ (see section \ref{sec:memory}) to compute reading and writing weights. %$\mathbf{M}^{(\rm ref)}_{ij}\sim\mathcal{N}(0,100)$.

For Table \ref{tab:zsre_seq}, the reference memory is constructed by encoding the prompt for each of the 1000 edits, and placing it in one row of $\mathbf{M}^{(\rm ref)}$. %When writing each fact, this results in [each fact is written to the corresponding slot in M, with writing key 1-hot, since we have K = num edits]

For Figure \ref{fig:batch} (b), the reference memory is constructed by encoding the first prompt for each of the 1000 unique facts (among the several rephrasings in the edit set which are written to memory) and placing it in a single row in $\mathbf{M}^{(\rm ref)}$. 
Thus, when querying memory with an encoded rephrased prompt $\mathbf{z}$ in Eq. \eqref{eq:gaussian_conv}, if $\mathbf{z}$ is closest to the row $k$ in $\mathbf{M}^{(\rm ref)}$ corresponding to the same fact, the key vector element $w_k$ will be largest for this element, and suppressed for other memory locations.
(We use $\alpha=10^{-3}$ to strongly suppress more distant encodings in the reference memory. Empirically, we found that that the nearest-neighbor encoding picked out by Eq. \eqref{eq:gaussian_conv} with small $\alpha$ is usually the encoded prompt for the same fact, with lower F1 scores occurring mainly in cases where the nearest-neighbor row in $\mathbf{M}^{(\rm ref)}$ corresponds to a different fact.)
We found that computing reading and writing weights as in \cite{pham2021generative}, $\mathbf{w}=\mathbf{z}(\mathbf{M}^{(\rm ref)})^{\dagger}$, was not as effective with rephrased facts (Figure \ref{fig:batch} (b) and Table \ref{table_forget_rephrase_appendix}) unless the number of rephrasings per fact was relatively large.

%throughout, reading and writing weights from prompt
%Algo 2...

When writing to memory, a trailing period is appended to the ground truth label, in order to reduce the likelihood of the model generating additional text. When evaluating the F1 score, we remove (in both target and predicted tokens) the token corresponding to a period (13). We also remove the token 198, which corresponds to the new line character `\textbackslash n', when it is generated as the last token.

\begin{figure}
    \centering
    \includegraphics[width=0.9\linewidth]{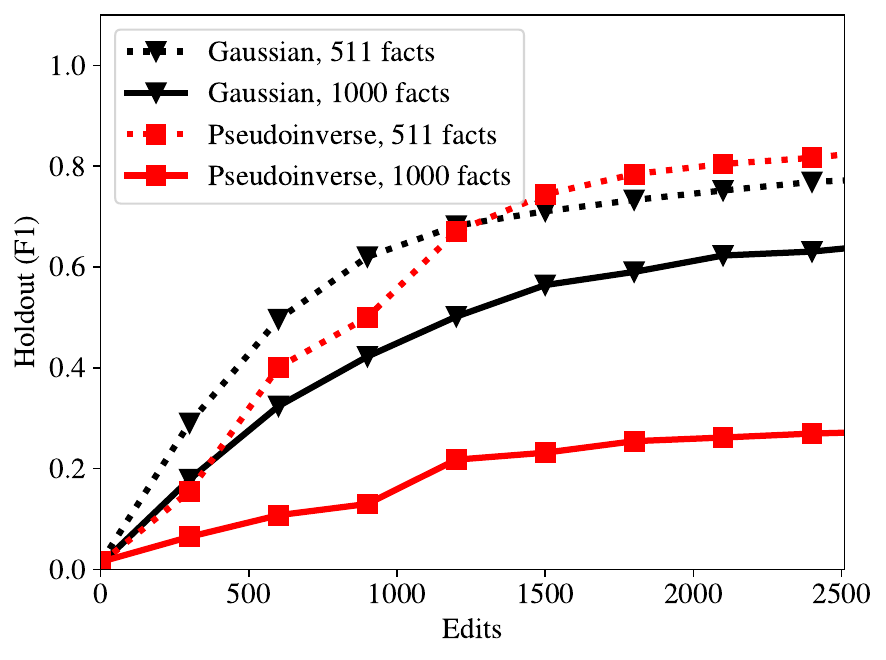}
    \caption{Mean F1 score of Larimar, comparing different choices for computing reading and writing weights -- the Gaussian convolution in Eq. \eqref{eq:gaussian_conv} and the pseudoinverse method of \cite{pham2021generative} -- on held-out sets of unseen rephrasings from ZsRE  over a sequence of 3000  edits. (Black curves are shown in Figure \ref{fig:batch} (b) in the main text.) }
    \label{fig:zsre_seq_gen_appendix}
\end{figure}

In Figure~\ref{fig:zsre_seq_gen_appendix}, we compare different variants of Larimar, on the same task as shown in Figure~\ref{fig:batch} (b). Relative to the Gaussian convolution method of Eq. \eqref{eq:gaussian_conv}, computing reading and writing weights with the reference memory matrix pseudoinverse, $\mathbf{w}=\mathbf{z}(\mathbf{M}^{(\rm ref)})^{\dagger}$ performed well on a dataset of 511 ZsRE facts and $\approx20$ phrasings per fact, but significantly worse on a dataset of $1000$ ZsRE with $10$ phrasings per fact. (We hypothesize that Eq. \eqref{eq:gaussian_conv} is more effective at finding a nearby rephrase encoding for the same fact when there are only one or a few paraphrases available in the data.)
%Lastly, we show a variant which uses Eq. \eqref{eq:gaussian_conv}, but instead of using the encoding of the prompt (e.g. ``What country is John Smith from?'') uses the encoding of the nouns in the prompt (e.g. ``John Smith country''), with proper nouns first. This performs $\sim5\%$ better, most likely due to a reduction of the variance between encodings for different rephrases of the same fact.
%(iii) %upper bound, encode answers (or fact IDs)

In our fact forgetting experiments (Table \ref{forgetting_counterfact}), we used a simple reference memory where each matrix element is sampled randomly, $\mathbf{M}^{(\rm ref)}_{ij}\sim\mathcal{N}(0,1)$.
We found this choice to be less effective when querying with rephrased prompts -- in which case the additional structure of $\mathbf{M}^{(\rm ref)}$ described above helps to locate the nearby encoding of a different phrasing of the same fact -- but to be sufficient when querying with the same prompts used when writing to memory (as in Table \ref{forgetting_counterfact}). % in which case the reading and writing keys are equivalent regardless of $\mathbf{M}^{(\rm ref)}$).} %(We observed comparable results for a broad range of Gaussian noise levels.) 
In this case we compute the writing weight using the encoding of the prompt of the fact written to memory, $\mathbf{W} = \mathbf{Z}_{\rm prompt} (\mathbf{M}^{(\rm ref)})^{\dagger}$ (instead of Eq. \eqref{eq:gaussian_conv}), and compute the reading weight in the same way, with the reading prompt differing from the writing prompt in rephrasing experiments.

Lastly, in our batch editing experiment (Figure \ref{fig:batch}), we computed writing weights using the encoded prompt, $\mathbf{W} = \mathbf{Z}_{\rm prompt} (\mathbf{M}^{(\rm ref)})^{\dagger}$, and computed both writing and reading weights with $\mathbf{M}^{(\rm ref)}$ set to the memory matrix obtained from Larimar's training (although we found a Gaussian random matrix to yield comparable results).

Throughout these experiments, we use $\sigma_w=0$ and $\xi=0$. %, which yielded marginally better performance than $\xi=O(10^{-4})$

\section{Generalization via Rephrase-Augmented Memory}

We also evaluate Larimar-1.3B on generalization to unseen rephrasings, by writing a variable number of seen rephrases of the same fact to memory.
%We also evaluate Larimar-1.3B on  generalization to unseen rephrasings of facts written in memory, while preserving the ability to forget specified facts.
After writing $N_{reph}$  rephrasings for each of $N_{fact}$ facts to memory, % forgetting is done  for all the written rephrasings of one specified fact. 
%We then 
we estimate  recall by querying the model with $N_{reph}$  unseen rephrasings. % of (i) the retained facts, and (ii) the forgotten fact. 
{\color{black}(As in the sequential editing experiment with rephrase queries, we use %the Gaussian convolution method (Appendix \ref{app:details}) to compute reading and writing weights, using 
a reference memory matrix constructed from the prompt encodings for the facts written to memory.)} 
In Table \ref{table_forget_rephrase_appendix}, we show average recall of the ground-truth answer %(defined as binary variable indicating the presence or absence of the exact ground truth answer in the model response) 
for samples from the ZsRE validation set, revealing generalization to unseen rephrases. % for both forgetting and read from memory.
Naturally, for facts with more rephrases in memory, recall is higher. % showing the ability of Larimar to generalize to unseen 
We furthermore compare the Gaussian convolution method of Eq. \eqref{eq:gaussian_conv} %, as in Table \ref{rephrase_forgetting}, 
to computing reading and writing weights with the reference memory matrix pseudoinverse, $\mathbf{w}=\mathbf{z}(\mathbf{M}^{(\rm ref)})^{\dagger}$. As in Figure \ref{fig:zsre_seq_gen_appendix}, Eq. \eqref{eq:gaussian_conv} leads to better recall with fewer rephrasings per fact, but falls short when there are many rephrasings per fact.

\iffalse
\begin{table}
\centering
\begin{center}
    \begin{adjustbox}{width=0.28\textwidth}    
    \begin{tabular}{ l  l  l  l l l}
    \hline
    ($N_{fact}$, $N_{reph}$) & Forgotten & Retained \\ \hline
    (20, 10) & 0.0 & \textbf{0.90} \\ 
    (40, 5) & 0.0 & 0.84 \\ 
    (100, 2) & 0.0 & 0.78 \\ 
    (200, 1) & 0.0 & 0.69 \\ 
    (1, 1) & 0.0 & 0.68 \\ \hline
    \end{tabular}
\end{adjustbox}    
\end{center}
\caption{Recall with rephrased facts, after writing $N_{reph}$ rephrasings for each of $N_{fact}$ ZsRE facts to Larimar-1.3B memory. {\color{black}For the single-fact case ($N_{fact}=1$) we report recall prior to the forgetting operation}.}
\label{rephrase_forgetting}
\end{table}
\fi

%In Table \ref{table_forget_rephrase_appendix}, we show additional results for the task of Table \ref{rephrase_forgetting} in the main text, this time omitting results for forgotten facts and reporting the mean recall for facts in memory.

\begin{table}
\centering
\begin{center}
    \begin{adjustbox}{width=0.36\textwidth}    
    \begin{tabular}{ l l l }
    \hline
    ($N_{fact}$, $N_{reph}$) & Pseudoinverse & Gaussian \\ \hline
    (20, 10) & \textbf{0.94} & 0.90 \\ 
    (40, 5) & \textbf{0.84} & \textbf{0.84} \\ 
    (100, 2) & 0.66 & \textbf{0.78} \\ 
    (200, 1) & 0.33 & \textbf{0.69} \\ 
    (1, 1) & 0.63 & \textbf{0.68} \\ \hline
    \end{tabular}
\end{adjustbox}    
\end{center}
\caption{Recall after writing $N_{reph}$ rephrasings for each of $N_{fact}$ ZsRE facts to Larimar-1.3B memory, and querying with unseen phrasings, using (i) $\mathbf{w}=\mathbf{z}(\mathbf{M}^{(\rm ref)})^{\dagger}$ (`pseudoinverse') or (ii) Eq. \eqref{eq:gaussian_conv}, `Gaussian.'}
\label{table_forget_rephrase_appendix}
\end{table}

\section{Generation Robustness}
We assess Larimar's robustness to sampling noise of the reading weights ($\sigma_w$) in terms of edit success and perplexity. To measure edit success, we use 2000 cases from the CounterFact dataset. For each case, the encoding of the prompt is concatenated with the 'new target' to form an episode, which is written to memory. Next we sample the weight vector $w \sim N(\bar{w},\sigma_w), w\in \mathbb{R}^K$ and take $z = w M$ to be the read-out vector, which is decoded along with the prompt. We then report the edit success.
To measure perplexity, we consider 1000 samples from the Wikipedia dataset. For each sentence, we write it into Larimar memory, and take the first 10 characters of the sentence as our prompt. We then perform generation as above. We repeat these steps for each of the 1000 sentences and then this text is fed into GPT2 large model to compute perplexity.

\begin{figure}
    \centering
    \includegraphics[width=0.62\linewidth]{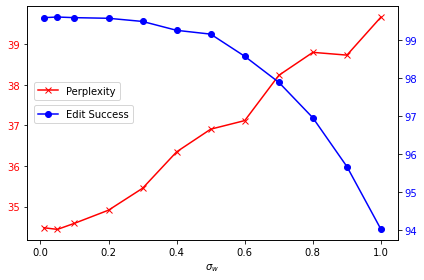}
    \caption{Generation perplexity and single fact edit success as a function of  varying magnitude of $\sigma_w$ for Larimar-6B. (Results show that our $Z_{readout}$ is robust to noise in the addressing/memory matrix and also leads to the correct response from the decoders) }
    \label{fig:ppl_rewrite}
\end{figure}

In Figure~\ref{fig:ppl_rewrite}, we report the perplexity and rewrite success metrics as a function of $\sigma_w,$ averaged over 3 independent runs. Overall the results indicate that Larimar is fairly robust to increased noise variance up to a range.  

\end{document}